\pdfoutput=1
\documentclass[11pt]{article}

\usepackage[final]{acl}

\usepackage{times}
\usepackage{latexsym}
\usepackage[most]{tcolorbox}

\usepackage[T1]{fontenc}
\usepackage[utf8]{inputenc}

\usepackage{microtype}

\usepackage{inconsolata}

\usepackage{graphicx}
\usepackage{graphicx}
\usepackage{times}
\usepackage{latexsym}
\usepackage{amsmath} % 引入amsmath宏包以提供各种数学功能
\usepackage{amssymb} % 引入amssymb宏包以使用\mathbb等命令
\usepackage{algorithm}
\usepackage{algpseudocode}
\usepackage{graphicx} 
\usepackage{tabularx}
\usepackage{booktabs}
\usepackage{multirow}
\usepackage{booktabs}
\usepackage{wrapfig}
\usepackage{caption}
\usepackage{enumitem}
\usepackage{tablefootnote}
\usepackage{array}
\usepackage{setspace}
\usepackage{CJKutf8} %中文编码
\usepackage{graphicx}
\usepackage{float}
\usepackage{subfigure}
\usepackage[T1]{fontenc}
\usepackage{graphicx}

\newcommand\blfootnote[1]{%
\begingroup
\renewcommand\thefootnote{}\footnote{#1}%
\addtocounter{footnote}{-1}%
\endgroup
}
\title{Memory Augmentation Unlocks Efficient Chain-of-Thought Reasoning}

\author{
    ~Simeng Zhang$^{1,2,*}$,
    ~Yilong Chen$^{1,2,*}$,
    ~Wenyuan Zhang$^{4}$\\
    ~\textbf{Zhenyu Zhang}$^{3}$,
    ~\textbf{Yao Chen}$^{1,2}$,
    ~\textbf{Junyuan Shang}$^{3}$,
    ~\textbf{Tingwen Liu}$^{1,2\dagger}$ \\
    \normalsize $^1$ Institute of Information Engineering, Chinese Academy of Sciences\\
    \normalsize $^2$ School of Cyber Security, University of Chinese Academy of Sciences\\
    \normalsize $^3$ Baidu Inc. \normalsize $^4$ Tencent Inc.\\
     \{\texttt{zhangsimeng, chenyilong, liutingwen\}@iie.ac.cn}
}

\begin{document}
\maketitle
\begin{abstract}
Large language models often rely on Chain-of-Thought (CoT) reasoning to solve complex tasks, but verbose reasoning traces introduce substantial inference overhead.
CoT compression shortens generation, yet aggressive compression may disrupt logical coherence and degrade performance.
We formalize this trade-off as the \textit{Context-Generation Substitution Law}, where explicit reasoning context substitutes for part of decode-time generation.
Based on this principle, we propose \textit{Memory-Augmented Compression}, a training-free framework that constructs reusable reasoning memories from historical traces and retrieves them as prefill-side scaffolds.
Rather than using raw demonstrations, these memories summarize reusable reasoning patterns, key constraints, and critical operations to compensate for information lost during compression.
Experiments show that Memory consistently improves prompt-based Chain-of-Draft (CoD) compression across mathematical reasoning, complex reasoning, and science question answering tasks, yielding accuracy gains of 21.4, 28.0, 29.5, and 6.61 points over CoD on GSM8K, MATH, BBH, and MMLU-Sci, while achieving a 1.14--1.49$\times$ latency speedup over standard CoT.
Memory is also compatible with token-level, reasoning-trace-level, and inference-state compression mechanisms.
Further analyzes show that the gains come from relevant reasoning memories rather than simply increasing context length.
\end{abstract}

\blfootnote{\hspace*{-1.8em}%
$^*$ Equal contribution. $^\dagger$ Corresponding author.\\An initial version of this work was completed in January 2026.
}

\section{Introduction}

% --- Paragraph 1: Background (System 1 vs. System 2) ---
Large Language Models (LLMs) exhibit strong generalization capabilities, enabling them to perform complex reasoning tasks. Chain-of-Thought (CoT) prompting~\citep{wei2022chain} improves reasoning by explicitly generating intermediate steps, while recent reasoning-oriented models such as OpenAI o-series models~\citep{jaech2024openai} and DeepSeek-R1~\citep{guo2025deepseek} further show that scaling test-time computation and producing longer reasoning traces can substantially improve performance on difficult benchmarks. However, this long-thinking paradigm comes with a significant cost: verbose reasoning traces must be generated token by token in an autoregressive manner, leading to increased decoding latency, token cost, and serving overhead~\citep{kwon2023efficient,chen2026towards}.

\begin{figure}[t]  
\centering  
\includegraphics[width=8.0cm]{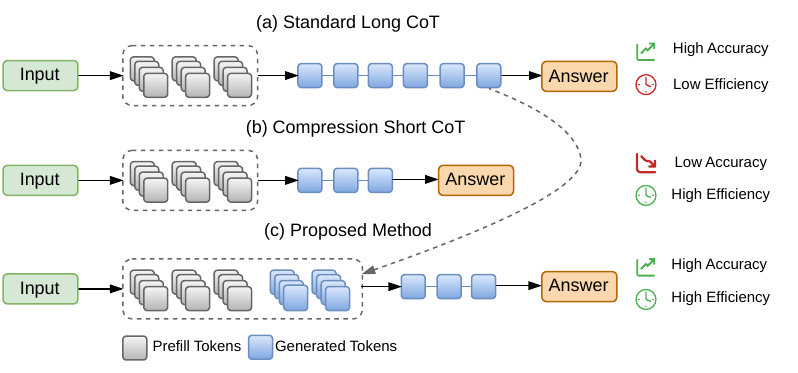}
\caption{
Illustration of memory-augmented compression.
Explicit memory shifts part of the reasoning support to the prefill stage, improving short-CoT accuracy while preserving generation efficiency.
}
\label{fig:motivation}
\end{figure}

% --- Paragraph 2: The Gap (Efficiency Bottleneck & Hardware Asymmetry) ---
To mitigate the overhead of long reasoning traces, existing methods compress reasoning at different levels, including semantic-level shortening, token-level pruning or skipping, and inference-state compression~\citep{xu2025chain,xia2025tokenskip,song2026reasoning}. However, these methods mainly reduce reasoning information produced or retained during generation. As compression becomes more aggressive, they may discard useful subgoals, constraints, or logical dependencies, causing accuracy degradation. This exposes a key limitation of generation-side compression: shorter reasoning traces do not necessarily preserve the support needed for problem solving. How, then, can we preserve such support without returning to long autoregressive generation?

% \vspace{0.5em}
\textit{How can we leverage the parallelism of context scaling to substitute the serial overhead of lengthy generation?}
% \vspace{0.5em}

% --- Paragraph 3: Theoretical Insight (Memory Compensation Effect - No Symbols, No Bold) ---
Motivated by this question, we conduct a preliminary study on the interplay between input context and generated CoT.
We observe that useful reasoning context, such as abstract reasoning memories, can help preserve or improve accuracy while enabling shorter reasoning traces.
Since prefill computation is more parallelizable than autoregressive decoding, this suggests that some decode-side reasoning support can be shifted to the prefill context.
We formalize this observation as the \textit{Context-Generation Substitution Law} in Sec.~\ref{sec:substitution-law}, capturing the trade-off between prefill-side reasoning context and decode-side CoT generation.

% --- Paragraph 4: Method & Contributions (Structured List - No Bold Stats) ---
Based on the above observation, we propose \textit{Memory-Augmented Compression}, a training-free memory-prefill framework for compensating information loss in compressed reasoning. 
The framework uses abstract reasoning memories constructed from historical traces, which capture reusable problem-solving patterns, key constraints, and critical operations. 
For a new query, relevant memories are retrieved and injected into the prompt as prefill-side scaffolds, allowing the model to access reusable reasoning support while generating shorter reasoning traces. 
The framework is not tied to a specific compression algorithm, making it compatible with various compression mechanisms, including prompting-based compression, token-level compression, and KV-cache compression.

Our contributions are:
\begin{itemize}
\item We introduce the \textit{Context-Generation Substitution Law}, the first formulation to characterizes the trade-off between prefill-side reasoning context and decode-time generation.

\item We propose \textit{Memory-Augmented Compression}, a training-free framework that retrieves abstract reasoning memories and injects them into the prefill context to facilitate highly compressed reasoning.

\item Experiments demonstrate that the proposed method improves the accuracy-latency trade-off across domains, models, and compression mechanisms, with analyses confirming that relevant memories compensate for information lost under aggressive compression.

\end{itemize}

\section{Preliminaries}
\label{sec:preliminaries}

\subsection{Standard Chain-of-Thought Inference}
We formalize the standard reasoning process of an LLM $\mathcal{M}_\theta$. 
Let $\mathbf{x} = (\mathcal{I}, q)$ denote the concatenation of the system instruction $\mathcal{I}$ and the user query $q$. 
Instead of directly mapping $\mathbf{x}$ to the final answer $y$, CoT inference first generates an intermediate reasoning sequence $z$~\cite{wei2022chain}. 
The joint probability can be decomposed as:
\begin{equation}
    P_\theta(y, z \mid \mathbf{x}) 
    = 
    P_\theta(z \mid \mathbf{x}) \cdot P_\theta(y \mid \mathbf{x}, z).
\end{equation}
Here, $P_\theta(z \mid \mathbf{x})$ corresponds to the reasoning phase, and $P_\theta(y \mid \mathbf{x}, z)$ represents the answering phase conditioned on the generated reasoning chain. 
While $z$ can improve answer accuracy, producing a long reasoning sequence incurs substantial autoregressive decoding cost.

\begin{figure*}[t]  
\centering  
\includegraphics[width=\textwidth]{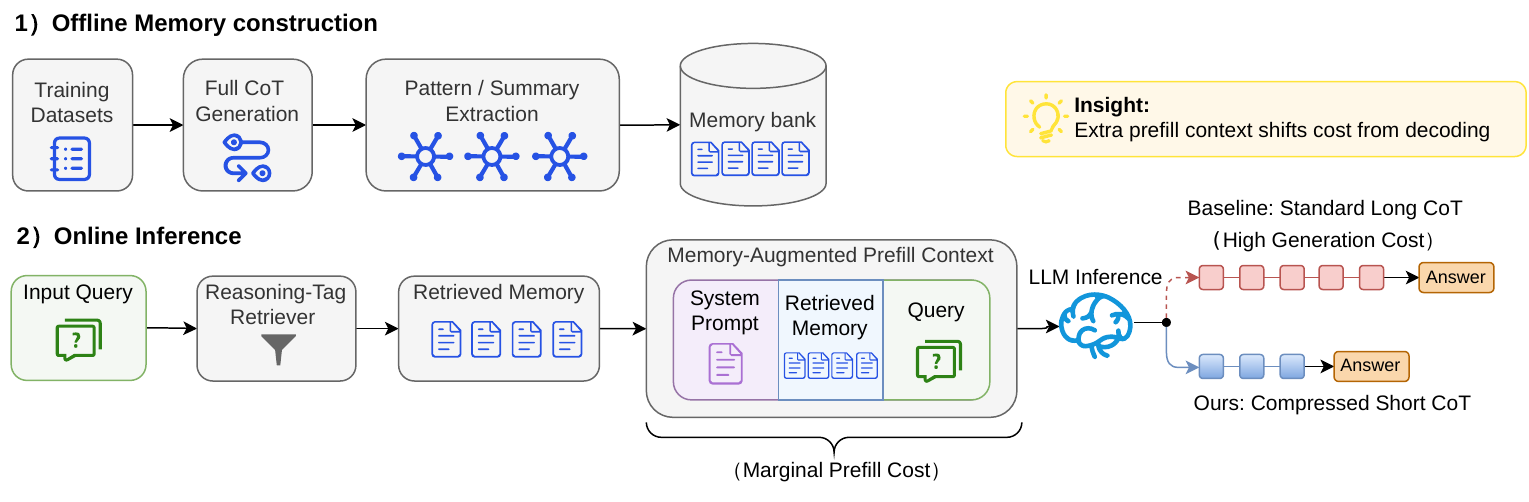}
\caption{
Overview of \textit{Memory-Augmented Compression}.
Historical reasoning traces are distilled into reusable memory entries offline.
At inference time, relevant memories are retrieved and injected into the prefill context to support compressed Short-CoT reasoning.
}
\label{fig:main_framework}
\end{figure*}

\subsection{The Inference Latency Bottleneck}
The computational cost of autoregressive decoding grows with the length of the generated sequence. 
Let $\mathcal{L}_{\mathrm{dec}} = |z|$ denote the decoding length of the reasoning chain.
In modern reasoning models, $\mathcal{L}_{\mathrm{dec}}$ can extend to thousands of tokens~\cite{guo2025deepseek,chen2026towards}, making the reasoning trace much longer than the final answer. 
Because tokens in $z$ are decoded sequentially, long reasoning traces create a latency bottleneck in LLM serving~\cite{kwon2023efficient}.
Although decoding acceleration methods such as speculative decoding improve generation throughput~\cite{leviathan2023fast}, they still require producing the reasoning trace. 
Therefore, reducing decode-time reasoning length remains critical for efficient reasoning.

\subsection{Context-Generation Substitution Law}
\label{sec:substitution-law}
To address the decoding bottleneck, we introduce an external memory source $\mathcal{C}$, which stores reusable reasoning information distilled from historical examples. 
Our core hypothesis is that relevant reasoning context can substitute for part of the reasoning trace otherwise produced through decode-time generation.
Given an input $\mathbf{x}$, we retrieve a structured reasoning memory
\begin{equation}
    M = \phi(\mathbf{x}, \mathcal{C}),
\end{equation}
and condition the model on $M$ instead of relying solely on a fully generated reasoning chain $z$:
\begin{equation}
    P_\theta(y \mid \mathbf{x}, M)
    \approx
    P_\theta(y \mid \mathbf{x}, z).
\end{equation}
Conditioned on $M$, the model produces a compressed reasoning chain (Short-CoT) $z'$, where ideally $|z'| \ll |z|$ while task performance is preserved.
For clarity, we focus on the LLM prefill--decode trade-off after memory retrieval and exclude retrieval cost from our latency formulation. We characterize the efficiency--accuracy trade-off as:
\begin{equation}
    \min_{\phi,\, z'} \mathcal{J}
    =
    |z'| + \gamma |M| + \lambda \mathcal{L}_{\mathrm{perf}},
\end{equation}
where $|z'|$ is the compressed decode-time reasoning length, $|M|$ is the added memory context length, and $\mathcal{L}_{\mathrm{perf}}$ denotes the performance penalty induced by compression. 
The coefficient $\gamma=\tau_{\mathrm{pre}}/\tau_{\mathrm{dec}}$ captures the relative cost between prefill processing and autoregressive decoding. 
Since prefill processing is more parallelizable than token-by-token decoding, adding moderate reasoning memory can be beneficial when it reduces decode length or improves compressed reasoning accuracy.
This formulation defines the \textit{Context-Generation Substitution Law}: useful reasoning context can substitute for part of the generated reasoning trace when its prefill cost is outweighed by savings or accuracy gains in compressed decoding.

\section{Memory-Augmented Compression}

\subsection{Cognitive Memory in LLM Reasoning}
\label{subsec:cognitive_memory_view}

We conceptualize LLM reasoning through a cognitive memory view, where different information sources incur different inference costs.

\noindent\textbf{Implicit memory.}
Pre-trained weights \(\theta\) store latent knowledge and reasoning ability, which are accessed through inference computation.

\noindent\textbf{Explicit memory.}
The input context provides directly accessible information, such as instructions, retrieved memories, or few-shot examples.
Since context tokens are processed in parallel during prefill, explicit memory can support reasoning at lower marginal cost than autoregressive generation.

\noindent\textbf{Working memory.}
The generated CoT sequence and its KV cache form dynamic working memory.
Because it is constructed token by token through autoregressive decoding, verbose working memory becomes the main inference bottleneck.

From this perspective, standard CoT reasoning externalizes information from implicit memory into working memory through decode-time generation.
Our goal is to move part of the reusable reasoning information into explicit memory in advance, reducing the amount of working memory the model needs to generate from scratch.

\subsection{Empirical Observations and Analysis}
\label{subsec:empirical_observation}
To validate the feasibility of substituting decode-time generation with explicit context, we conduct a preliminary study on the interaction between CoT compression and memory augmentation.

\begin{figure}[t]
    \centering
    \includegraphics[width=\columnwidth]{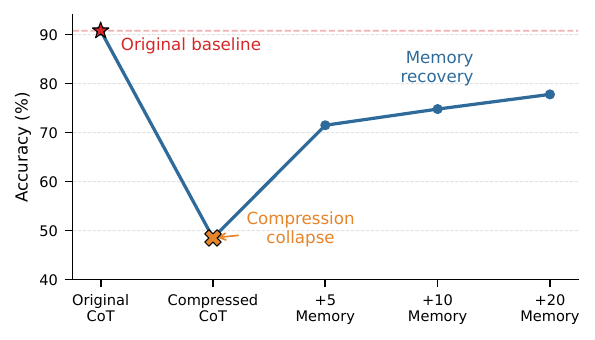}
    \caption{Explicit memory mitigates the reasoning collapse caused by aggressive compression.}
    \label{fig:observation}
\end{figure}

\noindent\textbf{Observation 1: Compression collapse.}
As shown in Figure~\ref{fig:observation}, aggressively reducing decode-time reasoning length leads to severe accuracy degradation, confirming that generated reasoning provides essential scaffolds for complex reasoning.

\noindent\textbf{Observation 2: Memory compensation.}
Injecting prior reasoning examples into the prompt substantially recovers the degradation caused by compression, suggesting that explicit memory can compensate for missing reasoning information.

\noindent\textbf{Computational trade-off analysis.}
Following the Context-Generation Substitution Law in Section~\ref{sec:substitution-law}, we model inference latency $\mathcal{T}$ under the hardware asymmetry between prefill and autoregressive decoding:
\begin{equation}
    \mathcal{T} \approx |\mathbf{x}| \cdot \tau_{\mathrm{pre}} + |z| \cdot \tau_{\mathrm{dec}},
\end{equation}
where $\mathbf{x}$ denotes the prefill context and $z$ denotes the generated reasoning trace.
Since autoregressive decoding proceeds sequentially, we typically have $\tau_{\mathrm{dec}} \gg \tau_{\mathrm{pre}}$.
Injecting relevant reasoning memory into the input context introduces an additional prefill cost $\Delta_{\mathrm{in}}$, but it can reduce the required decode-time reasoning length by $\Delta_{\mathrm{out}}$. 
The injected memory acts as a reasoning scaffold, allowing the model to bypass redundant or recoverable intermediate steps. 
A net latency reduction is achieved when:
\begin{equation}
    \underbrace{\Delta_{\mathrm{out}} \cdot \tau_{\mathrm{dec}}}_{\text{Decoding Gain}} 
    >
    \underbrace{\Delta_{\mathrm{in}} \cdot \tau_{\mathrm{pre}}}_{\text{Prefill Cost}}.
\end{equation}
Rearranging yields the efficiency condition:
\begin{equation}
    \frac{\Delta_{\mathrm{out}}}{\Delta_{\mathrm{in}}} 
    >
    \frac{\tau_{\mathrm{pre}}}{\tau_{\mathrm{dec}}}
    = \gamma.
\end{equation}
This condition indicates that explicit memory can provide a favorable trade-off when a small amount of added prefill context sufficiently reduces decode-time reasoning or recovers accuracy under compression. 
Empirically, our results show that Memory improves the accuracy--latency balance of compressed inference.

%%% 主实验表格
\begin{table*}[t]
\centering
\small
\setlength{\tabcolsep}{4pt}
\renewcommand{\arraystretch}{1.08}

\begin{tabular*}{\textwidth}{@{\extracolsep{\fill}} ll l r r r r r r}
\toprule \multicolumn{9}{c}{\textbf{(a) Across-domain CoD results with Qwen2.5-7B}} \\ \midrule
Domain & Dataset & Method & Acc. & $\Delta$ vs CoD & Prefill & Decode & Total & Latency  (ms)\\
\midrule
\multirow{6}{*}{Math}
& \multirow{3}{*}{GSM8K}
& CoT     & 91.4 & --    & 96.3   & 299.2 & 395.5  & 3742.5{\scriptsize\,(1.00$\times$)}\\
& & CoD     & 67.9 & --    & 112.3  & 51.1  & 163.4  & 666.6 {\scriptsize(5.61$\times$)} \\
& & +Memory & 89.3 & \textbf{+21.4} & 1161.9 & 187.2 & 1349.1 & 2515.4 {\scriptsize(1.49$\times$)} \\
\cmidrule(lr){2-9}
& \multirow{3}{*}{MATH}
& CoT     & 72.8 & --    & 105.9  & 569.3 & 675.2  & 5117.0 {\scriptsize(1.00$\times$)} \\
& & CoD     & 43.0 & --    & 121.9  & 100.3 & 222.2  & 1119.5 {\scriptsize(4.57$\times$)} \\
& & +Memory & 71.0 & \textbf{+28.0} & 1772.6 & 460.2 & 2232.8 & 4489.6 {\scriptsize(1.14$\times$)} \\
\midrule
\multirow{3}{*}{Complex}
& \multirow{3}{*}{BBH}
& CoT     & 60.8 & --    & 134.0 & 320.9 & 454.9  & 4007.1 {\scriptsize(1.00$\times$)} \\
& & CoD     & 41.0 & --    & 150.0 & 46.3  & 196.3  & 652.1 {\scriptsize(6.14$\times$)} \\
& & +Memory & \textbf{70.5} & \textbf{+29.5} & 983.0 & 143.3 & 1126.3 & 3148.5 {\scriptsize(1.27$\times$)} \\
\midrule
\multirow{3}{*}{Science}
& \multirow{3}{*}{MMLU-Sci}
& CoT     & 62.11 & --    & 119.0  & 452.5 & 571.5  & 4821.0 {\scriptsize(1.00$\times$)} \\
& & CoD     & 60.35 & --    & 135.0  & 48.1  & 183.1  & 572.6 {\scriptsize(8.42$\times$)} \\
& & +Memory & \textbf{66.96} & \textbf{+6.61} & 1504.9 & 276.9 & 1781.8 & 3423.2 {\scriptsize(1.41$\times$)} \\
\bottomrule
\end{tabular*}

\vspace{0.6em}

\begin{tabular*}{\textwidth}{@{\extracolsep{\fill}} ll l l l r r r}
\toprule
\multicolumn{8}{c}{\textbf{(b) Memory as a plug-in for compression methods}} \\
\midrule
Method & Model & Dataset & Compression & Memory Config & Base Acc & +Memory Acc & Gain \\
\midrule
\multirow{4}{*}{TokenSkip}
& Qwen2.5-7B & GSM8K & $\gamma=0.1$ & $k=3$, Long-CoT & 52.62 & 70.13 & +17.51 \\
& Qwen2.5-7B & GSM8K & $\gamma=0.4$ & $k=3$, Long-CoT & 68.08 & 76.88 & +8.80 \\
& LLaMA-3.1-8B & GSM8K & $\gamma=0.9$ & $k=5$, Long-CoT & 77.10 & 79.23 & +2.13 \\
\midrule
\multirow{3}{*}{RPC}
& DS-R1-Qwen-7B & AIME 2024 & $P=4096$ & $k=1$, Summary & 52.92 & 55.00 & +2.08 \\
& DS-R1-Qwen-7B & AIME 2024 & $P=2048$ & $k=1$, Summary & 51.25 & 53.33 & +2.08 \\
& DS-R1-Qwen-7B & AIME 2024 & $P=1024$ & $k=1$, Summary & 47.50 & 51.67 & +4.17 \\
\midrule
\multirow{2}{*}{Extra-CoT}
& Qwen2.5-7B & MATH & $\gamma=0.8$  & $k=5$, Long-CoT & 39.80 & 41.20 & +1.40 \\
& Qwen2.5-7B & MATH & $\gamma=1.0$ & $k=5$, Long-CoT & 35.80 & 40.40 & +4.60 \\
\bottomrule
\end{tabular*}

\caption{
Main results of \textit{Memory-Augmented Compression}.
Panel (a) reports across-domain CoD results with Qwen2.5-7B.
Panel (b) evaluates Memory as a plug-in for representative compression settings.
}
\label{tab:main_results}
\end{table*}

% --- [Part 3] pipeline
\subsection{Framework Design}
\label{subsec:method_pipeline}

Motivated by the cognitive memory view and empirical observations above, we propose \textit{Memory-Augmented Compression}, a plug-in framework that shifts reasoning support from decode-time generation to prefill-side memory. As shown in Figure~\ref{fig:main_framework}, it consists of four components: memory bank construction, memory retrieval, memory-augmented prefill, and memory-guided compressed inference.

\paragraph{Memory bank.}
We construct a memory bank $\mathcal{B}$ from previously solved reasoning
examples. Unlike conventional few-shot prompting, MAC does not directly
inject raw input--output demonstrations into the inference context.
Instead, each example is transformed into a memory representation that
captures reusable reasoning information, such as the problem type, key
constraints, subgoals, solution strategy, and critical operations.
Only the selected memory content is serialized and injected at inference
time, serving as external reasoning guidance for compressed reasoning.
Details of the memory representation and construction procedure are
provided in Appendix~\ref{app:memory_representation} and
Appendix~\ref{app:memory_construction}.

\paragraph{Memory retrieval.}
Given a query $x$, the retriever selects a set of relevant memories from $\mathcal{B}$, with the retrieval size controlled by $k$. 
In our implementation, retrieval is guided by reasoning tags and semantic similarity, so that the selected memories share similar problem-solving structures with the query rather than surface wording alone:
\begin{equation}
M_x = \mathcal{R}(x, \mathcal{B}, k).
\label{eq:memory_retrieval}
\end{equation}

\paragraph{Memory prefill and Inference.}
The retrieved memories are concatenated with the system prompt and query to form a memory-augmented prefill context.
Conditioned on this context, the model generates a concise reasoning trace and final answer.
Since the memories encode abstract reasoning structures rather than full demonstrations, they provide compact scaffolds that help preserve key subgoals, constraints, and critical operations under aggressive compression.

\section{Experiments}
\label{sec:experiments}

\subsection{Experimental Setup}
\paragraph{Datasets.}
We evaluate on diverse datasets across domains, including GSM8K~\cite{cobbe2021training}, MATH~\cite{hendrycks2021measuring}, BBH~\cite{suzgun2023challenging}, MMLU-Sci~\cite{hendrycks2020measuring}, and AIME 2024.

\paragraph{Models.}
Unless otherwise specified, we use LLaMA-3.1-8B and Qwen2.5-7B as open-weight backbones~\cite{grattafiori2024llama,yang2024qwen2}, and additionally evaluate API-based reasoning models for cross-model generalization.

\paragraph{Baselines.}
We compare standard CoT~\cite{wei2022chain}, Chain-of-Draft (CoD)~\cite{xu2025chain} as the prompt-based compression baseline, and memory-augmented variants.
To evaluate plug-in ability, we further combine Memory with TokenSkip~\cite{xia2025tokenskip}, RPC~\cite{song2026reasoning}, and Extra-CoT~\cite{tang2026towards}.

\paragraph{Memory and metrics.}
For each dataset, we instantiate the memory bank using previously solved examples and exclude all evaluation samples to prevent data leakage.
At inference time, we retrieve the top-$k$ relevant memories and inject them as prefill-side scaffolds without parameter updates.
We report accuracy, prefill tokens, decode tokens, total tokens, and latency.
For memory-augmented methods, $\Delta$Acc denotes the accuracy gain over the corresponding compressed baseline, while ``+Memory'' denotes applying \textit{Memory-Augmented Compression} to that baseline.

\subsection{Main Results}
\label{sec:main_results}

Table~\ref{tab:main_results} evaluates Memory from two perspectives: the
accuracy--latency trade-off under CoD and compatibility with other compression
mechanisms.
In Panel~(a), Latency measures LLM inference time (excluding retrieval), with
parentheses indicating speedup over standard CoT; retrieval overhead is
reported separately in Table~\ref{tab:retriever}.
In Panel~(b), Base and +Memory use the same setting except for memory
injection; full results are provided in Appendix~\ref{app:plugin_results}.

\paragraph{General evaluation across domains.}
As shown in Table~\ref{tab:main_results}(a), CoD greatly shortens decode-time reasoning but often causes large accuracy drops, indicating that aggressive compression removes useful intermediate reasoning information.
Adding Memory consistently recovers performance, improving over CoD by 21.4, 28.0, 29.5, and 6.61 points on GSM8K, MATH, BBH, and MMLU-Sci, respectively.
CoD+Memory approaches standard CoT on GSM8K and MATH and surpasses it on BBH and MMLU-Sci, suggesting that retrieved memories provide targeted reasoning scaffolds rather than merely restoring compressed steps.

\paragraph{Accuracy--latency trade-off.}
Although Memory increases prefill tokens, it reduces decode-side reasoning
compared with standard CoT and achieves a 1.14--1.49$\times$ latency speedup
across domains.
When the retrieval overhead is added back, MAC still remains faster than CoT
end-to-end across datasets (full breakdown in Appendix).
Thus, Memory improves accuracy over the compressed CoD baseline while preserving MAC’s end-to-end speed advantage, providing empirical support for the \textit{Context--Generation Substitution Law}.

\paragraph{Plug-in compatibility.}
Table~\ref{tab:main_results}(b) shows that memory augmentation can be combined with compression mechanisms beyond CoD.
For each row, Base and +Memory use the same model, dataset, compression setting, and decoding setup, with memory injection as the only difference.
Memory augmentation improves TokenSkip across different compression ratios and models, and also yields positive gains with RPC and Extra-CoT.
Notably, these methods include both training-based approaches, such as TokenSkip and Extra-CoT, and the training-free RPC method, while memory augmentation requires no additional training.
These results suggest that the memory module serves as a plug-in compensation mechanism for recovering reasoning information lost under different efficiency methods.

%% 消融实验1:检索【表】
\begin{table}[t]
\centering
\small
\setlength{\tabcolsep}{4pt}
\renewcommand{\arraystretch}{1.08}
\begin{tabular*}{\columnwidth}{@{\extracolsep{\fill}}lcc}
\toprule
Retriever & Accuracy (\%) & Overhead \\
\midrule
No Memory (CoD)  & 43.00 & -- \\
Random  & 50.20 & 0.1\,ms \\
Query Emb.  & 53.20 & 100\,ms \\
BM25 & 54.60 & 1\,ms \\
\textbf{Reasoning Tag} & \textbf{56.20} & 683\,ms \\
\bottomrule
\end{tabular*}
\caption{Accuracy and online retrieval overhead of different retrieval
strategies on MATH-500. Overhead is the approximate per-query online cost.}
\label{tab:retriever}
\end{table}

%% 消融实验2:内容【表】
\begin{table}[t]
\centering
\small
\setlength{\tabcolsep}{4pt}
\renewcommand{\arraystretch}{1.08}
\begin{tabular*}{\columnwidth}{@{\extracolsep{\fill}}lrrrr}
\toprule
Memory Content & Acc. & Prefill & Decode & $\Delta$Acc \\
\midrule
CoD (None) & 55.00 & 136.31 & 2105.50 & 0.00 \\
Few-shot & 46.67 & 300.79 & 1406.11 & -8.33 \\
Summary & \textbf{61.67} & 606.81 & 2117.04 & \textbf{+6.67} \\
Short-CoT & 46.67 & 605.40 & 1454.78 & -8.33 \\
Long-CoT & 43.33 & 961.37 & 1531.02 & -11.67 \\
\bottomrule
\end{tabular*}
\caption{
Impact of memory representation on DeepSeek-V3.2/AIME 2024 with top-$k=1$ retrieval and 2048 maximum generation tokens.
}
\label{tab:memory_representation}
\end{table}

%%% 消融实验
\subsection{Ablation Study}
\label{subsec:ablation}

To dissect the contribution of each component in our Memory-Augmented paradigm, we conduct component-wise ablation studies.

%% 消融实验1:检索策略
\paragraph{Impact of retrieval strategy.}
We compare Query Embedding, BM25, and Reasoning Tag to examine the
accuracy--latency trade-off of retrieval (Table~\ref{tab:retriever}).
Reasoning Tag achieves the highest accuracy and is therefore used as the
default retriever, despite the additional cost of online tag generation
($\approx$683\,ms/query). BM25 provides a faster alternative
($\approx$1\,ms) while retaining most of the accuracy. Importantly, all
retrievers substantially outperform the no-memory CoD baseline,
indicating that MAC is robust to the choice of retriever and can be
adapted to different deployment constraints.

%% 消融实验2:格式
\paragraph{Impact of memory representation.}
We ablate whether the injected content should be fixed few-shot demonstrations, truncated/full CoT traces, or abstract reasoning memories.
Table~\ref{tab:memory_representation} shows that fixed few-shot demonstrations and CoT-trace memories degrade performance, while Summary Memory yields a +6.67 point gain.
This suggests that Memory benefits from compact reasoning abstractions rather than simply adding demonstrations or longer reasoning traces.

%% 消融实验3:K值
\paragraph{Impact of memory size.}
Table~\ref{tab:memory_size} shows how memory size $k$ affects accuracy and
prefill cost on Qwen2.5-7B. Accuracy improves from $k{=}1$ to the best $k$, but pushing $k$ to 20 brings no gain and even
hurts accuracy while raising prefill cost, revealing a coverage--cost
trade-off. The larger drop on the harder MATH benchmark suggests over-injection
is more damaging for complex reasoning.

%% 消融实验3: K值【表】
\begin{table}[t]
\centering
\small
\setlength{\tabcolsep}{3pt}
\renewcommand{\arraystretch}{1.08}
\begin{tabular*}{\columnwidth}{@{\extracolsep{\fill}}lrrrrrrr}
\toprule
& \multicolumn{2}{c}{$k{=}1$}
& \multicolumn{3}{c}{Best}
& \multicolumn{2}{c}{$k{=}20$} \\
\cmidrule(lr){2-3}
\cmidrule(lr){4-6}
\cmidrule(lr){7-8}
Dataset
& Acc. & Pre.
& $k$ & Acc. & Pre.
& Acc. & Pre. \\
\midrule
GSM8K
& 73.39 & 259
& 14 & 80.52 & 1881
& 79.68 & 2628 \\

MATH
& 52.20 & 349
& 16 & 62.60 & 3330
& 57.40 & 4120 \\
\bottomrule
\end{tabular*}

\caption{Effect of memory size $k$ on Qwen2.5-7B. We report accuracy
(Acc.) and prefill tokens (Pre.). Accuracy improves from $k{=}1$ to the best
$k$, whereas $k{=}20$ may reduce accuracy and increases prefill cost.}
\label{tab:memory_size}
\end{table}

%% 分析实验
\subsection{Analysis}
\label{subsec:analysis}

To provide a comprehensive understanding of the properties and boundaries of our framework, we performed a series of analytical experiments.

%% 分析实验1:推理模型
\paragraph{Scalability to reasoning models.}
We evaluate whether \textit{Memory-Augmented Compression} scales to stronger
API-based reasoning models on AIME 2024.
As shown in Table~\ref{tab:reasoning_models}, CoD+Memory consistently improves
over CoD across DeepSeek-V3.2, Qwen3.5-plus, and o4-mini, with gains of +6.67,
+9.17, and +4.17 points.
This suggests that explicit memory provides model-agnostic reasoning support
beyond a specific open-weight backbone.
Although Memory increases prefill tokens, decode tokens remain close to CoD,
indicating that the gains mainly come from prefill-side support rather than
longer generation.

%% 分析实验1:推理模型【表】
\begin{table}[t]
\centering
\small
\setlength{\tabcolsep}{3.5pt}
\renewcommand{\arraystretch}{1.08}
\begin{tabular*}{\columnwidth}{@{\extracolsep{\fill}}lrrrrr}
\toprule
Method & Acc. (\%) & Prefill & Decode & Total & $\Delta$Acc \\
\midrule
\multicolumn{6}{l}{\textit{DeepSeek-V3.2}} \\
CoT     & 55.00 & 124.01 & 2733.94 & 2857.95 & -- \\
CoD     & 55.00 & 136.31 & 2105.50 & 2241.81 & 0.00 \\
+Memory & \textbf{61.67} & 606.81 & 2117.04 & 2723.85 & \textbf{+6.67} \\
\midrule
\multicolumn{6}{l}{\textit{Qwen3.5-plus}} \\
CoT     & 75.83 & 142.30 & 8657.60 & 8799.90 & -- \\
CoD     & 49.17 & 157.80 & 7533.00 & 7690.80 & -26.66 \\
+Memory & \textbf{85.00} & 666.20 & 7364.90 & 8031.10 & \textbf{+9.17} \\
\midrule
\multicolumn{6}{l}{\textit{o4-mini}} \\
CoT     & 74.17 & 133.33 & 4547.59 & 4680.93 & -- \\
CoD     & 70.83 & 143.72 & 3900.59 & 4044.31 & -3.34 \\
+Memory & \textbf{75.00} & 738.23 & 4031.10 & 4769.33 & \textbf{+4.17} \\
\bottomrule
\end{tabular*}
\caption{
Scalability to API-based reasoning models on AIME 2024. +Memory denotes CoD+Memory. All results are reported as pass@8, following common practice on AIME. Gains are in percentage points over CoD.
}
\label{tab:reasoning_models}
\end{table}

%% 分析实验2: Prefill--Decode
\begin{table}[t]
\centering
\small
\setlength{\tabcolsep}{3.5pt}
\renewcommand{\arraystretch}{1.08}
\begin{tabular*}{\columnwidth}{@{\extracolsep{\fill}}lrrrr}
\toprule
Dataset & Batch & Prefill & Decode & Ratio \\
        &       & (ms/tok) & (ms/tok) & $\tau_{\text{dec}}/\tau_{\text{pre}}$ \\
\midrule
GSM8K    & 1 & 0.136 & 11.81 & 87.0 \\
GSM8K    & 8 & 0.144 & 2.11  & 14.7 \\
MATH-500 & 1 & 0.150 & 11.97 & 79.4 \\
MATH-500 & 8 & 0.134 & 2.96  & 22.1 \\
\bottomrule
\end{tabular*}
\caption{
Per-token prefill and decode latency for Qwen2.5-7B on an NVIDIA H20
under batch sizes.
}
\label{tab:cost}
\end{table}

%% 分析实验3:不同压缩率【图】
\begin{figure}[t]
    \centering
    \includegraphics[width=\linewidth]{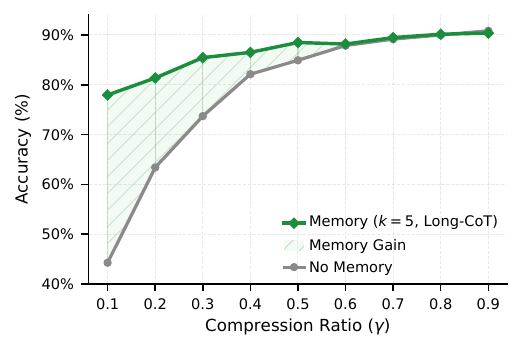}
    \caption{Memory compensation across compression ratios on GSM8K with
Qwen2.5-7B. Smaller $\gamma$ indicates more aggressive compression. Green:
memory-augmented ($k{=}5$); gray: no-memory; shaded: gain.}
    \label{fig:memory_compensation_ratio}
\end{figure}

%% 分析实验2:Prefill--decode
\paragraph{Prefill--decode cost quantification.}
MAC replaces part of expensive autoregressive generation with comparatively
inexpensive prefill context. We compute the average per-token prefill and
decode latency, denoted by $\tau_{\mathrm{pre}}$ and
$\tau_{\mathrm{dec}}$, over the full evaluation set.
As shown in Table~\ref{tab:cost}, one decode token incurs the same latency
as approximately $14.7$--$87.0$ prefill tokens, depending on the dataset
and batch size. More generally, injecting $M$ memory tokens while eliminating
$D$ decode tokens reduces latency when
\begin{equation}
    M\tau_{\mathrm{pre}} < D\tau_{\mathrm{dec}}.
\end{equation}
The advantage remains substantial at batch size $8$: although batching
amortizes autoregressive decoding and narrows the prefill--decode gap, it
does not eliminate the computational asymmetry.

%% 分析实验3:不同压缩率
\paragraph{Memory compensation under different compression ratios.}
We analyze how memory compensation varies with the compression ratio $\gamma$
on GSM8K using Qwen2.5-7B, where a smaller $\gamma$ denotes more aggressive
compression. As shown in Figure~\ref{fig:memory_compensation_ratio}, the gain
is strongly compression-dependent: it is largest under aggressive compression,
where the no-memory baseline drops sharply, and diminishes as the baseline
recovers under milder compression. This is consistent with memory compensating
for reasoning information lost by compression, rather than providing a constant
gain. A full sweep over retrieval size $k{\in}\{3,5,10\}$ and both CoT formats
on MATH (Appendix) shows the same trend, confirming robustness to $k$.

\begin{figure*}[t]
    \centering
    \includegraphics[width=15.8cm]{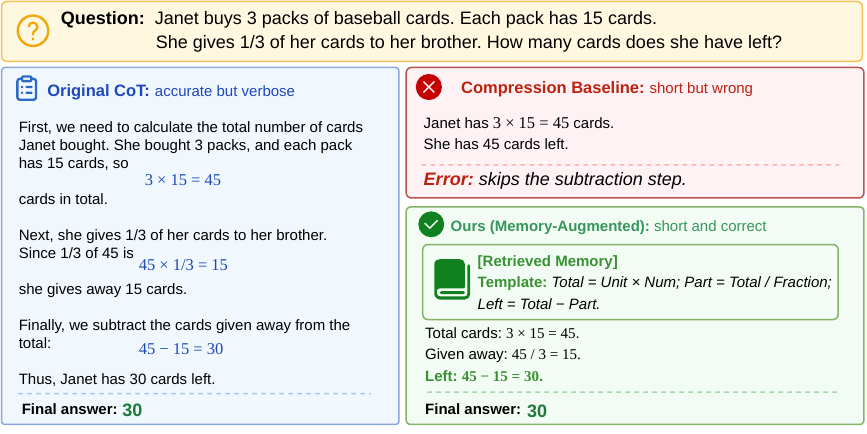}
    \caption{Case study on GSM8K where retrieved memory guides compressed reasoning to recover the missing subtraction step.}
    \label{fig:case study}
\end{figure*}

%% 分析实验4:case study
\paragraph{Case study.}
Figure~\ref{fig:case study} presents a qualitative GSM8K example.
The compressed baseline retains the total-card computation but drops the subsequent subtraction step, incorrectly treating the total as the final answer.
In contrast, the retrieved memory provides a structural template of \textit{Total}, \textit{Part}, and \textit{Left}, guiding the compressed reasoning to preserve the critical ``Left = Total - Part'' operation.
This illustrates that memory supplies missing reasoning structure, helping prevent logic collapse under aggressive compression.

%% 分析实验4:case study

\section{Related Work}
\label{sec:related_work}

\subsection{Context Augmentation and Memory}
Prior work improves inference by augmenting or optimizing the input context,
such as retrieving in-context demonstrations~\citep{liu2022makes} or
constructing examples with rationales~\citep{zhang2022automatic}.
Long-context methods~\citep{jin2024llm,zhang2025long} and context compression
methods~\citep{xu2023recomp,jiang2023llmlingua,jiang2024longllmlingua}
increase context capacity or reduce processing cost.
More closely related, reasoning-memory methods reuse prior reasoning to reduce
redundant computation~\citep{yang2024buffer,didolkar2025metacognitive,ahmed2025retrieval}.
MAC instead uses retrieved reasoning as a pluggable compensation layer for
intermediate content omitted by CoT compression, while decoupling the memory,
retriever, and compressor. This design shifts part of the reasoning burden from
decode-time generation to prefill context and allows each component to be
replaced independently.

\subsection{CoT Compression and Efficient Reasoning}
Recent work shortens CoT through pruning or token skipping~\citep{hou2025thinkprune,xia2025tokenskip,fan2026ctrlcot,tang2026towards}, budget-aware or self-compressed reasoning~\citep{lin2025plan,deng2026conpress,cao2026draft}, and reinforcement learning for concise reasoning~\citep{han2026long,tian2026shorter,qiao2025concise}.
These methods act on generation, but aggressive compression may remove
information needed for correct reasoning.
MAC instead injects retrieved reasoning scaffolds before generation,
providing a pluggable memory layer compatible with different compressors.

\subsection{System-Level Generation Acceleration}
Another line of work reduces inference cost through speculative decoding and its variants~\citep{leviathan2023fast,cai2024medusa,li2024eagle,fu2024break}, model sparsification~\citep{frantar2023sparsegpt}, and KV-cache compression~\citep{ramachandran2025thinkv}.
These methods make a given output cheaper or faster to generate, whereas MAC reduces the amount of explicit decode-time reasoning required by supplying part of the reasoning support through the prefill context.
The two directions are orthogonal in mechanism and can, in principle, be combined.

\section{Conclusion}
\label{sec:conclusion}

We introduced \textit{Memory-Augmented Compression}, a training-free framework that improves compressed reasoning by retrieving reusable reasoning memories as prefill-side scaffolds.
Instead of only shortening decode-time generation, our method shifts part of the reasoning support to the input context, compensating for information lost under compression.
Experiments across diverse reasoning tasks show that Memory improves accuracy over compressed baselines such as CoD while achieving lower latency than standard CoT.
It is also compatible with multiple efficiency mechanisms, including prompt-based, token-level, and inference-state compression.
Further analyses show that the gains come from relevant reasoning memories and are most pronounced under aggressive compression, supporting explicit memory as a promising direction for efficient reasoning.

\section*{Limitations}
In this section, we discuss the potential limitations and risks of our work. First, API-based models do not provide
a consistent interface for measuring model-internal prefill and decode
latency, which limits direct runtime comparisons across models. Second,
MAC introduces additional overhead from query-tag generation, embedding,
retrieval, and memory prefilling, and its latency advantage may therefore
be smaller for queries requiring only short reasoning traces. Finally,
MAC depends on the quality of the memory bank and retrieval results.
Irrelevant or poorly matched memories may introduce distracting
information and degrade reasoning quality. Our experiments mainly use
verified historical examples, so robustness to noisier memories remains
to be further evaluated.

% \section*{Acknowledgments}

\bibliography{custom}

@article{wei2022chain,
  title={Chain-of-thought prompting elicits reasoning in large language models},
  author={Wei, Jason and Wang, Xuezhi and Schuurmans, Dale and Bosma, Maarten and Xia, Fei and Chi, Ed and Le, Quoc V and Zhou, Denny and others},
  journal={Advances in neural information processing systems},
  volume={35},
  pages={24824--24837},
  year={2022}
}

@article{jaech2024openai,
  title={Openai o1 system card},
  author={Jaech, Aaron and Kalai, Adam and Lerer, Adam and Richardson, Adam and El-Kishky, Ahmed and Low, Aiden and Helyar, Alec and Madry, Aleksander and Beutel, Alex and Carney, Alex and others},
  journal={arXiv preprint arXiv:2412.16720},
  year={2024}
}

@article{guo2025deepseek,
  title={Deepseek-r1: Incentivizing reasoning capability in llms via reinforcement learning},
  author={Guo, Daya and Yang, Dejian and Zhang, Haowei and Song, Junxiao and Wang, Peiyi and Zhu, Qihao and Xu, Runxin and Zhang, Ruoyu and Ma, Shirong and Bi, Xiao and others},
  journal={arXiv preprint arXiv:2501.12948},
  year={2025}
}

@inproceedings{kwon2023efficient,
  title={Efficient memory management for large language model serving with pagedattention},
  author={Kwon, Woosuk and Li, Zhuohan and Zhuang, Siyuan and Sheng, Ying and Zheng, Lianmin and Yu, Cody Hao and Gonzalez, Joseph and Zhang, Hao and Stoica, Ion},
  booktitle={Proceedings of the 29th symposium on operating systems principles},
  pages={611--626},
  year={2023}
}

@article{chen2026towards,
  title={Towards reasoning era: A survey of long chain-of-thought for reasoning large language models},
  author={Chen, Qiguang and Qin, Libo and Liu, Jinhao and Peng, Dengyun and Guan, Jiannan and Wang, Peng and Hu, Mengkang and Zhou, Yuhang and Gao, Te and Che, Wanxiang},
  journal={Science China Information Sciences},
  volume={69},
  number={6},
  pages={161101},
  year={2026},
  publisher={Springer}
}

@article{xu2025chain,
  title={Chain of draft: Thinking faster by writing less},
  author={Xu, Silei and Xie, Wenhao and Zhao, Lingxiao and He, Pengcheng},
  journal={arXiv preprint arXiv:2502.18600},
  year={2025}
}

@inproceedings{xia2025tokenskip,
  title={Tokenskip: Controllable chain-of-thought compression in llms},
  author={Xia, Heming and Leong, Chak Tou and Wang, Wenjie and Li, Yongqi and Li, Wenjie},
  booktitle={Proceedings of the 2025 Conference on Empirical Methods in Natural Language Processing},
  pages={3351--3363},
  year={2025}
}

@article{song2026reasoning,
  title={Reasoning path compression: Compressing generation trajectories for efficient llm reasoning},
  author={Song, Jiwon and Jo, Dongwon and Kim, Yulhwa and others},
  journal={Advances in Neural Information Processing Systems},
  volume={38},
  pages={139724--139755},
  year={2026}
}

@inproceedings{leviathan2023fast,
  title={Fast inference from transformers via speculative decoding},
  author={Leviathan, Yaniv and Kalman, Matan and Matias, Yossi},
  booktitle={International Conference on Machine Learning},
  pages={19274--19286},
  year={2023},
  organization={PMLR}
}

@article{cobbe2021training,
  title={Training verifiers to solve math word problems},
  author={Cobbe, Karl and Kosaraju, Vineet and Bavarian, Mohammad and Chen, Mark and Jun, Heewoo and Kaiser, Lukasz and Plappert, Matthias and Tworek, Jerry and Hilton, Jacob and Nakano, Reiichiro and others},
  journal={arXiv preprint arXiv:2110.14168},
  year={2021}
}

@article{hendrycks2021measuring,
  title={Measuring mathematical problem solving with the math dataset},
  author={Hendrycks, Dan and Burns, Collin and Kadavath, Saurav and Arora, Akul and Basart, Steven and Tang, Eric and Song, Dawn and Steinhardt, Jacob},
  journal={arXiv preprint arXiv:2103.03874},
  year={2021}
}

@inproceedings{suzgun2023challenging,
  title={Challenging big-bench tasks and whether chain-of-thought can solve them},
  author={Suzgun, Mirac and Scales, Nathan and Sch{\"a}rli, Nathanael and Gehrmann, Sebastian and Tay, Yi and Chung, Hyung Won and Chowdhery, Aakanksha and Le, Quoc and Chi, Ed and Zhou, Denny and others},
  booktitle={Findings of the Association for Computational Linguistics: ACL 2023},
  pages={13003--13051},
  year={2023}
}

@article{hendrycks2020measuring,
  title={Measuring massive multitask language understanding},
  author={Hendrycks, Dan and Burns, Collin and Basart, Steven and Zou, Andy and Mazeika, Mantas and Song, Dawn and Steinhardt, Jacob},
  journal={arXiv preprint arXiv:2009.03300},
  year={2020}
}

@article{grattafiori2024llama,
  title={The llama 3 herd of models},
  author={Grattafiori, Aaron and Dubey, Abhimanyu and Jauhri, Abhinav and Pandey, Abhinav and Kadian, Abhishek and Al-Dahle, Ahmad and Letman, Aiesha and Mathur, Akhil and Schelten, Alan and Vaughan, Alex and others},
  journal={arXiv preprint arXiv:2407.21783},
  year={2024}
}

@article{yang2024qwen2,
  title={Qwen2. 5-math technical report: Toward mathematical expert model via self-improvement},
  author={Yang, An and Zhang, Beichen and Hui, Binyuan and Gao, Bofei and Yu, Bowen and Li, Chengpeng and Liu, Dayiheng and Tu, Jianhong and Zhou, Jingren and Lin, Junyang and others},
  journal={arXiv preprint arXiv:2409.12122},
  year={2024}
}

@article{tang2026towards,
  title={Towards efficient large language reasoning models via extreme-ratio chain-of-thought compression},
  author={Tang, Yuntian and Jia, Bohan and Huang, Wenxuan and Zhang, Lianyue and Xie, Jiao and Li, Wenxi and Ji, Rongrong and Lin, Shaohui},
  journal={arXiv preprint arXiv:2602.08324},
  year={2026}
}

@article{jin2024llm,
  title={Llm maybe longlm: Self-extend llm context window without tuning},
  author={Jin, Hongye and Han, Xiaotian and Yang, Jingfeng and Jiang, Zhimeng and Liu, Zirui and Chang, Chia-Yuan and Chen, Huiyuan and Hu, Xia},
  journal={arXiv preprint arXiv:2401.01325},
  year={2024}
}

@inproceedings{zhang2025long,
  title={Long context compression with activation beacon},
  author={Zhang, Peitian and Liu, Zheng and Xiao, Shitao and Shao, Ninglu and Ye, Qiwei and Dou, Zhicheng},
  booktitle={International Conference on Learning Representations},
  volume={2025},
  pages={101932--101948},
  year={2025}
}

@article{xu2023recomp,
  title={Recomp: Improving retrieval-augmented lms with compression and selective augmentation},
  author={Xu, Fangyuan and Shi, Weijia and Choi, Eunsol},
  journal={arXiv preprint arXiv:2310.04408},
  year={2023}
}

@inproceedings{jiang2023llmlingua,
  title={Llmlingua: Compressing prompts for accelerated inference of large language models},
  author={Jiang, Huiqiang and Wu, Qianhui and Lin, Chin-Yew and Yang, Yuqing and Qiu, Lili},
  booktitle={Proceedings of the 2023 conference on empirical methods in natural language processing},
  pages={13358--13376},
  year={2023}
}

@inproceedings{jiang2024longllmlingua,
  title={Longllmlingua: Accelerating and enhancing llms in long context scenarios via prompt compression},
  author={Jiang, Huiqiang and Wu, Qianhui and Luo, Xufang and Li, Dongsheng and Lin, Chin-Yew and Yang, Yuqing and Qiu, Lili},
  booktitle={Proceedings of the 62nd Annual Meeting of the Association for Computational Linguistics (Volume 1: Long Papers)},
  pages={1658--1677},
  year={2024}
}

@article{hou2025thinkprune,
  title={Thinkprune: Pruning long chain-of-thought of llms via reinforcement learning},
  author={Hou, Bairu and Zhang, Yang and Ji, Jiabao and Liu, Yujian and Qian, Kaizhi and Andreas, Jacob and Chang, Shiyu},
  journal={arXiv preprint arXiv:2504.01296},
  year={2025}
}

@article{fan2026ctrlcot,
  title={Ctrlcot: Dual-granularity chain-of-thought compression for controllable reasoning},
  author={Fan, Zhenxuan and Cao, Jie and Dai, Yang and Lv, Zheqi and Zhang, Wenqiao and Xie, Zhongle and Lu, Peng and Ooi, Beng Chin},
  journal={arXiv preprint arXiv:2601.20467},
  year={2026}
}

@article{lin2025plan,
  title={Plan and budget: Effective and efficient test-time scaling on large language model reasoning},
  author={Lin, Junhong and Zeng, Xinyue and Zhu, Jie and Wang, Song and Shun, Julian and Wu, Jun and Zhou, Dawei},
  journal={arXiv preprint arXiv:2505.16122},
  year={2025}
}

@article{deng2026conpress,
  title={ConPress: Learning Efficient Reasoning from Multi-Question Contextual Pressure},
  author={Deng, Jie and Liang, Shining and Li, Jun and Li, Hongzhi and Xie, Yutao},
  journal={arXiv preprint arXiv:2602.01472},
  year={2026}
}

@article{cao2026draft,
  title={Draft-Thinking: Learning Efficient Reasoning in Long Chain-of-Thought LLMs},
  author={Cao, Jie and Lin, Tianwei and Fan, Zhenxuan and Yuan, Bo and Zhao, Ziyuan and Yan, Rolan and Zhang, Wenqiao and Tang, Siliang},
  journal={arXiv preprint arXiv:2603.00578},
  year={2026}
}

@inproceedings{han2026long,
  title={Long Chain-Of-Thought Compression via Fine-Grained Group Policy Optimization},
  author={Han, Xinchen and Afifi, Hossam and Marot, Michel and Wang, Xilu and Yin, Lu},
  booktitle={ICASSP 2026-2026 IEEE International Conference on Acoustics, Speech and Signal Processing (ICASSP)},
  pages={4471--4475},
  year={2026},
  organization={IEEE}
}

@article{tian2026shorter,
  title={Shorter Thoughts, Same Answers: Difficulty-Scaled Segment-Wise RL for CoT Compression},
  author={Tian, Ye and Liu, Aijun},
  journal={arXiv preprint arXiv:2603.07598},
  year={2026}
}

@inproceedings{qiao2025concise,
  title={Concise: Confidence-guided compression in step-by-step efficient reasoning},
  author={Qiao, Ziqing and Deng, Yongheng and Zeng, Jiali and Wang, Dong and Wei, Lai and Wang, Guanbo and Meng, Fandong and Zhou, Jie and Ren, Ju and Zhang, Yaoxue},
  booktitle={Proceedings of the 2025 Conference on Empirical Methods in Natural Language Processing},
  pages={8021--8040},
  year={2025}
}

@article{li2024eagle,
  title={Eagle: Speculative sampling requires rethinking feature uncertainty},
  author={Li, Yuhui and Wei, Fangyun and Zhang, Chao and Zhang, Hongyang},
  journal={arXiv preprint arXiv:2401.15077},
  year={2024}
}

@article{fu2024break,
  title={Break the sequential dependency of llm inference using lookahead decoding},
  author={Fu, Yichao and Bailis, Peter and Stoica, Ion and Zhang, Hao},
  journal={arXiv preprint arXiv:2402.02057},
  year={2024}
}

@inproceedings{frantar2023sparsegpt,
  title={Sparsegpt: Massive language models can be accurately pruned in one-shot},
  author={Frantar, Elias and Alistarh, Dan},
  booktitle={International conference on machine learning},
  pages={10323--10337},
  year={2023},
  organization={PMLR}
}

@article{ramachandran2025thinkv,
  title={ThinKV: Thought-Adaptive KV Cache Compression for Efficient Reasoning Models},
  author={Ramachandran, Akshat and Neseem, Marina and Sakr, Charbel and Venkatesan, Rangharajan and Khailany, Brucek and Krishna, Tushar},
  journal={arXiv preprint arXiv:2510.01290},
  year={2025}
}

@article{cai2024medusa,
  title={Medusa: Simple llm inference acceleration framework with multiple decoding heads},
  author={Cai, Tianle and Li, Yuhong and Geng, Zhengyang and Peng, Hongwu and Lee, Jason D and Chen, Deming and Dao, Tri},
  journal={arXiv preprint arXiv:2401.10774},
  year={2024}
}

@inproceedings{liu2022makes,
  title={What makes good in-context examples for GPT-3?},
  author={Liu, Jiachang and Shen, Dinghan and Zhang, Yizhe and Dolan, William B and Carin, Lawrence and Chen, Weizhu},
  booktitle={Proceedings of Deep Learning Inside Out (DeeLIO 2022): The 3rd workshop on knowledge extraction and integration for deep learning architectures},
  pages={100--114},
  year={2022}
}

@article{zhang2022automatic,
  title={Automatic chain of thought prompting in large language models},
  author={Zhang, Zhuosheng and Zhang, Aston and Li, Mu and Smola, Alex},
  journal={arXiv preprint arXiv:2210.03493},
  year={2022}
}

@article{yang2024buffer,
  title={Buffer of thoughts: Thought-augmented reasoning with large language models},
  author={Yang, Ling and Yu, Zhaochen and Zhang, Tianjun and Cao, Shiyi and Xu, Minkai and Zhang, Wentao and Gonzalez, Joseph E and Cui, Bin},
  journal={Advances in Neural Information Processing Systems},
  volume={37},
  pages={113519--113544},
  year={2024}
}

@article{didolkar2025metacognitive,
  title={Metacognitive reuse: Turning recurring llm reasoning into concise behaviors},
  author={Didolkar, Aniket and Ballas, Nicolas and Arora, Sanjeev and Goyal, Anirudh},
  journal={arXiv preprint arXiv:2509.13237},
  year={2025}
}

@article{ahmed2025retrieval,
  title={Retrieval-of-Thought: Efficient Reasoning via Reusing Thoughts},
  author={Ahmed, Ammar and Khan, Azal Ahmad and Ahmad, Ayaan and Di, Sheng and Liu, Zirui and Anwar, Ali},
  journal={arXiv preprint arXiv:2509.21743},
  year={2025}
}
% \bibliography{main}

\section*{Appendix}

\appendix

\section{Method and Implementation Details}
\label{app:method}

This section provides additional implementation details of
\textit{Memory-Augmented Compression} (MAC).
We first describe the overall offline--online pipeline, followed by the
memory representation, memory-bank construction procedure, tag-based
retrieval, memory injection, and prompt templates used in our experiments.

\subsection{Overall Pipeline}
\label{app:pipeline}

MAC consists of two stages: offline memory-bank construction and online
memory-augmented compressed inference.

During the offline stage, we construct a reusable memory bank
$\mathcal{B}$ from historical solved examples
$\mathcal{D}_{\mathrm{mem}}$.
For each example, we collect or generate a valid reasoning trace and
associate it with structured reasoning tags or a summary representation.
The resulting memory bank is constructed once and reused across
evaluation queries.

During the online stage, given a new query $x$, MAC first constructs a
query-side retrieval representation and retrieves the top-$k$ relevant
memories from $\mathcal{B}$.
The retrieved memories are serialized and injected into the prefill
context together with the system instruction and the current query.
The backbone model then performs inference under a specified reasoning
compression method $\mathcal{A}$.

MAC does not update the parameters of the backbone model.
The memory representation, retriever, and compression method are
modular components and can be replaced independently.
Algorithm~\ref{alg:mac_pipeline} summarizes the overall procedure.

\begin{algorithm}[t]
\caption{\textsc{Memory-Augmented Compression}}
\label{alg:mac_pipeline}
\begin{algorithmic}[1]

\Require Historical examples $\mathcal{D}_{\mathrm{mem}}$,
query $x$, system instruction $\mathcal{I}$,
model $\mathcal{M}_{\theta}$, compression method $\mathcal{A}$,
memory format $f$, and retrieval size $k$

\Ensure Answer $y$ and compressed reasoning trace $z'$

\State $\mathcal{B}
\gets
\mathrm{BuildMemoryBank}(\mathcal{D}_{\mathrm{mem}}, f)$
\Comment{Performed once offline}

\Statex
\State \textbf{Online inference:}

\State $t_x \gets \mathrm{GenerateTags}(x)$

\State $\mathbf{v}_x \gets \mathrm{Embed}(t_x)$

\State $M_x
\gets
\mathrm{RetrieveTopK}(\mathbf{v}_x, \mathcal{B}, k)$

\State $\widetilde{M}_x
\gets
\mathrm{Serialize}(M_x)$

\State $\mathbf{x}_{\mathrm{mem}}
\gets
[\mathcal{I}; \widetilde{M}_x; x]$

\State $(z', y)
\gets
\mathcal{M}_{\theta}
(\mathbf{x}_{\mathrm{mem}}; \mathcal{A})$

\State \Return $(y, z')$

\end{algorithmic}
\end{algorithm}

\subsection{Memory Representation}
\label{app:memory_representation}

Each historical example is represented internally as a memory record
\begin{equation}
    e_i =
    (x_i,y_i,r_i,c_i,t_i,\mathbf{v}_i),
\end{equation}
where $x_i$ and $y_i$ denote the historical problem and its answer,
$r_i$ is the verified reasoning trace, $c_i$ is the memory content
constructed under the selected memory format, $t_i$ denotes the
structured reasoning tags, and $\mathbf{v}_i$ is the embedding of
$t_i$. The complete memory bank is defined as
\begin{equation}
    \mathcal{B}=\{e_i\}_{i=1}^{N}.
\end{equation}

The complete record $e_i$ is retained internally for offline
verification, retrieval, and analysis. At inference time, MAC injects
only the selected memory content $c_i$, rather than the complete record,
into the model context. Before injection, the memory content is
converted into a textual representation:
\begin{equation}
    m_i=\mathrm{Serialize}(c_i).
\end{equation}

We consider the following memory formats.

\paragraph{Long-CoT memory.}
The memory content $c_i$ contains the complete verified reasoning
trace, preserving detailed intermediate derivations, subgoals, and
operation sequences. Long-CoT memories retain the richest reasoning
information but incur the largest prefill cost.

\paragraph{Short-CoT memory.}
The memory content $c_i$ contains a condensed reasoning trace that
retains the principal solution steps and critical operations. This
format reduces the number of injected tokens while preserving more
procedural information than a high-level summary.

\paragraph{Summary memory.}
The memory content $c_i$ contains a structured summary of reusable
reasoning information, including the problem type, key constraints,
subgoals, solution strategy, and critical operations.

For a retrieved set
$M_x=\{e_{i_1},\ldots,e_{i_k}\}$, the corresponding injected memory
context is
\begin{equation}
    \widetilde{M}_x
    =
    [m_{i_1};m_{i_2};\ldots;m_{i_k}],
\end{equation}
where the memories are ordered according to their retrieval scores.

\subsection{Offline Memory-Bank Construction}
\label{app:memory_construction}

The memory bank is constructed exclusively from historical examples
that are disjoint from the corresponding evaluation set.
Algorithm~\ref{alg:memory_bank_construction} summarizes the offline
construction procedure, including reasoning-trace validation, memory
formatting, tag generation, and embedding computation.

\begin{algorithm}[t]
\caption{\textsc{Offline Memory Bank Construction}}
\label{alg:memory_bank_construction}
\begin{algorithmic}[1]

\Require Historical solved examples
$\mathcal{D}_{\mathrm{mem}}
=\{(x_i,y_i,r_i)\}_{i=1}^{N}$,
memory format
$f\in\{\text{Long-CoT},\text{Short-CoT},\text{Summary}\}$

\Ensure Memory bank $\mathcal{B}$

\State Initialize $\mathcal{B}\gets\emptyset$

\For{$(x_i,y_i,r_i)\in\mathcal{D}_{\mathrm{mem}}$}

    \If{$r_i$ is incorrect, incomplete, or unparsable}
        \State \textbf{continue}
    \EndIf

    \State $c_i\gets\mathrm{BuildMemory}(r_i,f)$

    \State $t_i\gets\mathrm{GenerateTags}(x_i,c_i)$

    \If{$\mathrm{FormatCheck}(c_i,t_i)$}
        \State $\mathbf{v}_i\gets\mathrm{Embed}(t_i)$
        \State $e_i\gets(x_i,y_i,c_i,t_i,\mathbf{v}_i)$
        \State $\mathcal{B}\gets\mathcal{B}\cup\{e_i\}$
    \EndIf

\EndFor

\State \Return $\mathcal{B}$

\end{algorithmic}
\end{algorithm}

\paragraph{Reasoning-trace collection.}
For each historical problem, we collect an available solution trace or
generate a full reasoning trace using the designated reasoning model.
The trace is paired with the corresponding answer and source metadata.

\paragraph{Correctness and completeness filtering.}
We retain only examples whose final prediction matches the reference
answer and whose reasoning trace contains a complete solution.
Examples with unparsable outputs, incomplete conclusions, or missing
solution annotations are discarded.

\paragraph{Memory annotation.}
For each retained example, we generate structured reasoning tags and,
when required, a condensed summary.
These annotations characterize the reusable reasoning structure of the
example rather than only its surface-level vocabulary.

\paragraph{Data separation.}
Each memory bank is constructed from historical examples that are
disjoint from the corresponding evaluation set.
Evaluation samples are never included in the memory bank or retrieval
pool, and their answers and reasoning traces are not used during
memory construction.
Depending on the evaluation setting, historical examples are drawn
from official training splits, official prompt exemplars, or historical
competition problems.
Dataset-specific memory sources and filtering procedures are detailed
in Appendix~\ref{app:datasets_memory}, with final bank sizes summarized
in Table~\ref{tab:memory_bank_summary}.

Once constructed, the memory bank remains fixed throughout evaluation.
No evaluation answer or evaluation reasoning trace is used to construct
or modify the memory bank.

\subsection{Tag Generation and Query Representation}
\label{app:tag_generation}

We use an LLM-based tag generator to produce structured reasoning
descriptions for historical memory examples.
The objective is to represent reusable reasoning attributes that may
not be captured by lexical similarity alone.

For each historical example, the tag generator receives the problem
together with the verified answer and produces a JSON object containing
the following fields:

\begin{itemize}
    \item \textbf{Knowledge Domains}: the primary mathematical,
    scientific, or logical domains involved in the problem.

    \item \textbf{Solution Strategies}: the main reasoning procedures,
    such as equation solving, case analysis, proportional reasoning,
    theorem application, or modeling.

    \item \textbf{Problem Type and Context}: the problem form and
    application context, such as an arithmetic word problem, geometric
    calculation, scientific question, or logical deduction.

    \item \textbf{Complexity}: the estimated reasoning depth, number of 
    required steps, degree of intermediate computation, and overall
    difficulty of solving the problem.
\end{itemize}

A simplified representation of the output schema is shown below.

\begin{tcolorbox}[
title=Reasoning-Tag Schema,
colback=white,
colframe=black!70,
colbacktitle=black!75,
coltitle=white,
fonttitle=\bfseries,
boxrule=0.5pt,
arc=2pt
]
\small\ttfamily
\{\\
\hspace*{1em}"knowledge\_domains": [...],\\
\hspace*{1em}"solution\_strategies": [...],\\
\hspace*{1em}"problem\_type\_and\_context": [...],\\
\hspace*{1em}"complexity": [...]\\
\}
\end{tcolorbox}

The generated tags are stored in $s_i$ and used to construct the
retrieval representation of each memory.

For an evaluation query $x$, the same tag schema is applied using only
the input question:
\begin{equation}
    s_x = \mathrm{TagGenerator}(x).
\end{equation}
The gold answer and gold reasoning trace of the evaluation query are
never provided to the tag generator.

Memory-side tags can be generated offline when the bank is constructed.
Query-side tags are generated from the current input before retrieval.
The exact tag-generation model and decoding settings are provided in
Appendix~\ref{app:decoding_hyperparameters}.

\subsection{Retrieval and Memory Injection}
\label{app:retrieval_injection}

Given a query $x$, we first obtain its structured tag representation
$s_x$ and encode it into a dense vector:
\begin{equation}
    \mathbf{q}_x = \mathrm{Enc}(s_x).
\end{equation}

For every memory entry $e_i$, the corresponding tag representation
$s_i$ is encoded as
\begin{equation}
    \mathbf{q}_i = \mathrm{Enc}(s_i).
\end{equation}

We compute the relevance score between the query and each memory using
cosine similarity:
\begin{equation}
    r_i =
    \frac{
        \mathbf{q}_x^{\top}\mathbf{q}_i
    }{
        \|\mathbf{q}_x\|_2
        \|\mathbf{q}_i\|_2
    }.
\end{equation}

The top-$k$ memories are selected according to
\begin{equation}
    M_x =
    \operatorname{TopK}_{e_i\in\mathcal{B}}(r_i,k).
\end{equation}

The selected memories are sorted in descending order of retrieval
score and serialized according to the selected memory format.
They are then concatenated with the system instruction and current
query:
\begin{equation}
    \mathbf{x}_{\mathrm{mem}}
    =
    [\mathcal{I};
    \widetilde{M}_x;
    x].
\end{equation}

The resulting sequence is processed during model prefill.
The model subsequently generates a compressed reasoning trace $z'$ and
final answer $y$ under compression method $\mathcal{A}$:
\begin{equation}
    (z',y)
    =
    \mathcal{M}_{\theta}
    (\mathbf{x}_{\mathrm{mem}};\mathcal{A}).
\end{equation}

MAC does not require parameter updates during retrieval or inference.
The retriever affects which memories are selected, while the compression
method independently determines how the model generates its
decode-time reasoning trace.

\subsection{Prompt Templates}
\label{app:prompts}

We provide the principal prompt templates used in our experiments.
The placeholder \texttt{\{question\}} is replaced by the current input
query, and \texttt{\{retrieved\_memory\}} is replaced by the serialized
top-$k$ memories.

\begin{tcolorbox}[
title=Standard CoT Prompt,
colback=white,
colframe=black!70,
colbacktitle=black!75,
coltitle=white,
fonttitle=\bfseries,
boxrule=0.5pt,
arc=2pt,
breakable
]
\small\ttfamily
Please reason step by step, and put your final answer within
\string\boxed\{\}.\\
Question: \{question\}
\end{tcolorbox}

\begin{tcolorbox}[
title=CoD Prompt,
colback=white,
colframe=black!70,
colbacktitle=black!75,
coltitle=white,
fonttitle=\bfseries,
boxrule=0.5pt,
arc=2pt,
breakable
]
\small\ttfamily
Think step by step, but only keep a minimum draft for each thinking
step, with 5 words at most. Put your final answer within
\string\boxed\{\}.\\
Question: \{question\}
\end{tcolorbox}

\begin{tcolorbox}[
title=Memory-Augmented Prompt,
colback=white,
colframe=black!70,
colbacktitle=black!75,
coltitle=white,
fonttitle=\bfseries,
boxrule=0.5pt,
arc=2pt,
breakable
]
\small\ttfamily
You are given relevant reasoning memories from historical examples.
Use them as external reasoning scaffolds, but solve the current question
independently.\\[0.3em]

[Retrieved Memories]\\
\{retrieved\_memory\}\\[0.3em]

[Current Question]\\
\{question\}\\[0.3em]

Think step by step, but keep the reasoning concise.
Put your final answer within \string\boxed\{\}.
\end{tcolorbox}

For memory-augmented variants, the retrieved-memory block is inserted
before the current query.
The evaluation query itself is unchanged, and the retrieved memories
provide auxiliary reasoning context rather than a replacement for
solving the current problem.

\section{Experimental Configuration}
\label{app:experimental_configuration}

This section provides the complete experimental configuration used in
our evaluation. We describe the dataset splits and memory sources,
backbone models and compression baselines, hardware and software
environment, decoding and retrieval hyperparameters, answer-evaluation
protocol, and latency measurement procedure.

\subsection{Datasets, Splits, and Memory Banks}
\label{app:datasets_memory}

We evaluate MAC on arithmetic, mathematical, symbolic, scientific, and
competition-level reasoning benchmarks, including
GSM8K~\citep{cobbe2021training},
MATH-500~\citep{hendrycks2021measuring},
BBH~\citep{suzgun2023challenging},
MMLU Science~\citep{hendrycks2020measuring},
and AIME 2024.
For every experimental setting, the memory source is disjoint from the
corresponding evaluation set.

Table~\ref{tab:memory_bank_summary} summarizes the evaluation splits,
memory sources, and final memory-bank sizes after correctness,
completeness, and format filtering.

\begin{table*}[t]
\centering
\small
\setlength{\tabcolsep}{5pt}
\renewcommand{\arraystretch}{1.08}
\begin{tabular*}{\textwidth}{
@{\extracolsep{\fill}}
llrlr
}
\toprule
Dataset
& Evaluation Split
& Eval. Size
& Memory Source
& Bank Size \\
\midrule
GSM8K
& Official test set
& 1,319
& Filtered official training examples
& 4,307
\\

MATH
& MATH-500
& 500
& Filtered official training examples
& 2,242
\\

BBH
& Nine selected tasks
& 495
& Official task-specific CoT exemplars
& 27
\\

MMLU-Sci
& Subject-stratified 20\% split
& 79
& Remaining 80\% from the same six subjects
& 317
\\

AIME 2024
& Official 2024 problems
& 30
& Historical AIME problems from 1983--2022
& 287
\\
\bottomrule
\end{tabular*}
\caption{
Dataset splits and memory-bank statistics.
Bank Size denotes the number of retained memories after correctness,
completeness, and format filtering.
Evaluation examples are excluded from all memory banks and retrieval
pools.
}
\label{tab:memory_bank_summary}
\end{table*}

\paragraph{GSM8K and MATH.}
For GSM8K and MATH, we construct memory banks from their official
training examples.
We retain only examples with a correct and complete reasoning trace.
GSM8K is evaluated on its official test set, while MATH is evaluated
on the MATH-500 subset.
The evaluation examples are not used during memory construction,
tag generation, or retrieval-index construction.

\paragraph{BBH.}
We evaluate nine BBH tasks and use the three official CoT exemplars
provided for each task as task-specific memories.
This produces 27 memories in total.
The exemplars are used only as historical reasoning support and remain
disjoint from the 495 evaluation examples.

\paragraph{MMLU Science.}
We select six science subjects:
college physics, high school physics, college chemistry, high school
chemistry, college biology, and high school biology.
We perform a subject-stratified 80/20 split, using 80\% of the examples
for memory construction and the remaining 20\% for evaluation.
This setting keeps the memory and evaluation distributions aligned
while ensuring that the two subsets remain disjoint.

\paragraph{AIME.}
For AIME 2024, we construct the memory bank from historical AIME
problems.
We begin with problems collected from 1983 to 2024 and exclude all
problems from 2023 and 2024 to prevent overlap with the evaluation
period.
We retain only problems with complete solution annotations and valid
answers.
The final bank contains 287 problems from 1983 to 2022.

\subsection{Models and Baselines}
\label{app:models_baselines}

\paragraph{Open-weight models.}
Our main open-weight experiments use Qwen2.5-7B and
LLaMA-3.1-8B.
We additionally evaluate Qwen2.5-72B to examine whether the effect of
memory-augmented compression persists at a larger model scale.
All models are used in their instruction-tuned form without additional
fine-tuning or parameter updates.

\paragraph{API-based reasoning models.}
We further evaluate MAC on three API-based reasoning models:
DeepSeek-V3.2, Qwen3.5-plus, and o4-mini.
For all models, we preserve the same memory-construction, retrieval,
and prompt-composition procedures whenever supported by the
corresponding API.
We report accuracy together with the numbers of prefill and decode
tokens, as well as their total, to characterize the effect of MAC on
reasoning performance and inference cost.

\paragraph{Reasoning baselines.}
Standard CoT prompts the model to generate an unconstrained
step-by-step reasoning trace.
Chain-of-Draft (CoD) constrains each intermediate reasoning step to at
most five words.
We additionally evaluate MAC as a plug-in module for TokenSkip, RPC,
and Extra-CoT.
For each compression method, the memory-augmented variant preserves the
original compression instruction and differs from its baseline only by
the addition of retrieved reasoning memories in the prefill context.

We use the officially released implementations of the baseline methods
whenever available.
When an official implementation cannot be directly integrated into our
evaluation framework, we reproduce the method according to the
configuration reported in the corresponding paper.

\subsection{Hardware and Software Environment}
\label{app:hardware_software}

Open-weight model inference is conducted on a server equipped with
eight NVIDIA H20 GPUs.
The server runs Ubuntu 22.04, Python 3.12, and CUDA 12.8.
Our implementation is based on PyTorch 2.5.1,
Transformers 4.52.4, and vLLM 0.6.4.post1.

We use vLLM for model serving, KV-cache management, and batched
generation.
For each backbone model, the tensor-parallel configuration remains
fixed across all compared methods.
Similarly, paired baseline and memory-augmented runs use the same
hardware, model checkpoint, numerical precision, batch size, and
serving configuration.

The memory bank, retrieval index, and memory-side tag embeddings are
constructed before evaluation.
No model parameters are updated during memory construction, retrieval,
or inference.

\subsection{Decoding and Hyperparameter Settings}
\label{app:decoding_hyperparameters}

Unless otherwise specified, open-weight models use deterministic greedy
decoding with temperature $T=0$ and $p_{\mathrm{top}}=1.0$.

Sampling is disabled for the main experiments.
For robustness experiments involving nonzero temperatures, the
corresponding temperature value is explicitly reported with the
results.

The maximum generation length is selected separately for standard and
compressed reasoning.
Standard CoT is assigned a sufficiently large output budget to avoid
truncating complete reasoning traces.
Compressed baselines and their memory-augmented counterparts use the
same maximum output length within each paired comparison.
Thus, adding memory does not provide an additional decode-token budget.

The retrieval size $k$, memory format, and compression strength are
specified for each experimental group. Unless otherwise indicated in the
corresponding table or figure, the same retrieval size and memory format
are used for all methods within a paired comparison. For retrieval-size
analyses, we vary $k\in\{1,3,5,8,10,12,14,16,18,20\}$. For memory-format
analyses, we compare Long-CoT, Short-CoT, and Summary memories under the
same retriever, backbone model, and compression configuration.

For TokenSkip experiments, the compression ratio is controlled by
$\gamma$, which is varied over
$\{0.1,0.2,0.3,0.4,0.5,0.6,0.7,0.8,0.9\}$.

For Extra-CoT, we evaluate the compression ratios reported in the
corresponding experiment.
All hyperparameters not directly related to memory injection are kept
unchanged between a compressed baseline and its memory-augmented
counterpart.

The tag generator uses deterministic decoding.
Memory-side tags and embeddings are generated once during offline
memory-bank construction.
For each evaluation input, query-side tags are generated using only the
current question and then encoded using the same embedding model as the
memory-side tags.
The gold answer and gold reasoning trace are never provided during
query-side tag generation.

\subsection{Evaluation Protocol}
\label{app:evaluation_protocol}

We report exact-match accuracy as the primary task-performance metric.
For each model output, we first extract the final answer from the
\texttt{\string\boxed\{\}} field when present.
If a valid boxed answer is unavailable, we apply the corresponding
benchmark-specific final-answer parser.

\paragraph{GSM8K.}
We normalize commas, whitespace, and numerical formatting before
comparing the predicted answer with the reference answer.
Only the final numerical answer is used for scoring.

\paragraph{MATH-500.}
We extract the final mathematical expression and evaluate it using the
same normalization and mathematical-equivalence procedure for all
methods.
Equivalent fractions, algebraic expressions, and normalized numerical
forms are treated as identical whenever recognized by the evaluator.

\paragraph{BBH and MMLU Science.}
For multiple-choice tasks, we extract the predicted option label and
compare it with the gold option.
Outputs that do not contain a valid option after parsing are counted as
incorrect.

\paragraph{AIME.}
AIME predictions are normalized to an integer in the range
$[0,999]$.
Leading zeros do not affect correctness.
Outputs that cannot be parsed as a valid integer are counted as
incorrect.

All methods are evaluated using the same answer-extraction and scoring
scripts.
Malformed, empty, or unparsable outputs are counted as incorrect rather
than discarded.
Accuracy is computed as
\begin{equation}
    \mathrm{Acc}
    =
    \frac{1}{|\mathcal{D}_{\mathrm{eval}}|}
    \sum_{i=1}^{|\mathcal{D}_{\mathrm{eval}}|}
    \mathbb{I}[\hat{y}_i=y_i].
\end{equation}

In addition to accuracy, we report average prefill tokens, average
decode tokens, total processed tokens, model latency, and end-to-end
latency when the required runtime measurements are available.

\subsection{Latency and Cost Measurement}
\label{app:latency_measurement}

We distinguish model execution latency from complete end-to-end
latency.
For an input query $x$, model latency is defined as
\begin{equation}
    T_{\mathrm{model}}
    =
    T_{\mathrm{prefill}}
    +
    T_{\mathrm{decode}}
\end{equation}
where $T_{\mathrm{prefill}}$ is the time required to process the complete
input context and $T_{\mathrm{decode}}$ is the time required for
autoregressive generation.

The end-to-end latency of MAC additionally includes online
query processing and retrieval:
\begin{equation}
\begin{aligned}
T_{\mathrm{e2e}}
&= T_{\mathrm{tag}}
 + T_{\mathrm{encode}}
 + T_{\mathrm{search}} \\
&\quad
 + T_{\mathrm{compose}}
 + T_{\mathrm{prefill}}
 + T_{\mathrm{decode}} .
\end{aligned}
\label{eq:e2e}
\end{equation} denotes memory serialization and prompt
construction.
Offline memory construction, memory-side tag generation, and
memory-index construction are excluded from per-query latency because
they are performed once and reused across evaluation inputs.

For prefill--decode cost analysis, we calculate the average per-token
latencies over the complete evaluation set:
\begin{equation}
    \tau_{\mathrm{pre}}
    =
    \frac{\sum_i T_{\mathrm{pre}}^{(i)}}
         {\sum_i N_{\mathrm{pre}}^{(i)}},
    \qquad
    \tau_{\mathrm{dec}}
    =
    \frac{\sum_i T_{\mathrm{dec}}^{(i)}}
         {\sum_i N_{\mathrm{dec}}^{(i)}}.
\end{equation}

All latency comparisons between a compressed baseline and its
memory-augmented counterpart are performed using the same model,
hardware, batch size, tensor-parallel configuration, and serving
process.
Unless otherwise stated, latency results are reported at batch size
one.
We additionally report batch-size-eight measurements to examine whether
the prefill--decode asymmetry persists under batched inference.

For API-based models, we calculate usage cost from the provider-reported
input and output token counts and the applicable token prices at the
time of evaluation.
Memory tokens are counted as input tokens, while generated reasoning
and final-answer tokens are counted as output tokens.
Because provider-side execution details are unavailable, API experiments
do not report separately measured prefill and decode latency.

\begin{table}[t]
\centering
\small
\setlength{\tabcolsep}{4pt}
\renewcommand{\arraystretch}{1.08}
\begin{tabular*}{\columnwidth}{
@{\extracolsep{\fill}}lrrrr
}
\toprule
Method
& Acc.
& Prefill
& Decode
& Latency (ms) \\
\midrule
CoT
& 77.00
& 105.9
& 597.4
& 22592.1
\\
CoD
& 62.20
& 121.9
& 122.9
& 4393.3
\\
CoD+Memory
& \textbf{78.20}
& 1772.6
& 373.0
& 13940.4
\\
\bottomrule
\end{tabular*}
\caption{
Results on MATH-500 with Qwen2.5-72B.
Prefill and Decode denote the average numbers of input and generated
tokens, respectively.
CoD+Memory achieves a $1.62\times$ speedup over full CoT while improving
accuracy from 77.00\% to 78.20\%.
}
\label{tab:qwen72b_results}
\end{table}

\begin{table}[t]
\centering
\small
\setlength{\tabcolsep}{4pt}
\renewcommand{\arraystretch}{1.05}
\begin{tabular*}{\columnwidth}{
@{\extracolsep{\fill}}lrrrr
}
\toprule
Model & $\gamma$ & Base & +Mem. & Gain \\
\midrule

\multirow{9}{*}{Qwen2.5-7B}
& 0.1 & 52.62 & 74.60 & +21.99 \\
& 0.2 & 60.65 & 74.53 & +13.87 \\
& 0.3 & 66.19 & 78.77 & +12.59 \\
& 0.4 & 68.08 & 80.74 & +12.66 \\
& 0.5 & 71.11 & 84.00 & +12.89 \\
& 0.6 & 77.26 & 84.99 & +7.73 \\
& 0.7 & 78.39 & 86.20 & +7.81 \\
& 0.8 & 82.79 & 86.88 & +4.09 \\
& 0.9 & 84.76 & 88.93 & +4.17 \\

\midrule

\multirow{9}{*}{LLaMA-3.1-8B}
& 0.1 & 52.99 & 79.08 & +26.08 \\
& 0.2 & 66.19 & 80.21 & +14.03 \\
& 0.3 & 73.46 & 80.74 & +7.28 \\
& 0.4 & 74.37 & 80.36 & +5.99 \\
& 0.5 & 77.26 & 81.12 & +3.87 \\
& 0.6 & 77.10 & 81.73 & +4.62 \\
& 0.7 & 79.38 & 81.27 & +1.90 \\
& 0.8 & 79.68 & 81.20 & +1.52 \\
& 0.9 & 80.82 & 82.56 & +1.74 \\

\bottomrule
\end{tabular*}
\caption{
Full TokenSkip plug-in results on GSM8K with top-$k=5$
Long-CoT memories.
}
\label{tab:tokenskip_gsm8k_full}
\end{table}

\begin{table}[t]
\centering
\small
\setlength{\tabcolsep}{4pt}
\renewcommand{\arraystretch}{1.05}
\begin{tabular*}{\columnwidth}{
@{\extracolsep{\fill}}lrrrr
}
\toprule
Model & $\gamma$ & Base & +Mem. & Gain \\
\midrule

\multirow{9}{*}{Qwen2.5-7B}
& 0.1 & 12.00 & 13.20 & +1.20 \\
& 0.2 & 19.60 & 22.20 & +2.60 \\
& 0.3 & 28.00 & 33.60 & +5.60 \\
& 0.4 & 34.40 & 44.80 & +10.40 \\
& 0.5 & 38.00 & 52.80 & +14.80 \\
& 0.6 & 42.60 & 58.40 & +15.80 \\
& 0.7 & 49.20 & 64.40 & +15.20 \\
& 0.8 & 50.40 & 67.60 & +17.20 \\
& 0.9 & 55.40 & 70.80 & +15.40 \\

\midrule

\multirow{9}{*}{LLaMA-3.1-8B}
& 0.1 & 18.00 & 7.20 & -10.80 \\
& 0.2 & 17.60 & 19.80 & +2.20 \\
& 0.3 & 14.20 & 30.60 & +16.40 \\
& 0.4 & 22.20 & 37.60 & +15.40 \\
& 0.5 & 24.60 & 41.60 & +17.00 \\
& 0.6 & 29.40 & 43.20 & +13.80 \\
& 0.7 & 31.80 & 43.40 & +11.60 \\
& 0.8 & 34.20 & 43.40 & +9.20 \\
& 0.9 & 37.20 & 43.00 & +5.80 \\

\bottomrule
\end{tabular*}
\caption{
Full TokenSkip plug-in results on MATH-500 with top-$k=5$
Long-CoT memories.
}
\label{tab:tokenskip_math_full}
\end{table}

\section{Additional Experimental Results}
\label{app:additional_results}

This section provides the complete experimental results omitted from
the main paper due to space constraints. We first report the overall
results across reasoning domains and compression methods, followed by
detailed analyses of compression plug-in compatibility, memory format,
retrieval size, compression strength, and decoding robustness.

\subsection{Model Scaling}
\label{app:scaling_api}

\paragraph{Scaling to Qwen2.5-72B.}
We further evaluate MAC on Qwen2.5-72B using MATH-500 to examine
whether the benefits of memory-augmented compression persist at a
larger model scale.
As shown in Table~\ref{tab:qwen72b_results}, CoD substantially reduces
decode length and latency relative to full CoT, but incurs a notable
accuracy drop.
Adding retrieved reasoning memories recovers and slightly surpasses the
full-CoT accuracy, while remaining considerably faster than full CoT.

Compared with full CoT, CoD reduces average decode tokens from 597.4 to
122.9 and lowers latency from 22592.1\,ms to 4393.3\,ms, but its
accuracy decreases by 14.8 percentage points.
CoD+Memory increases the prefill context to 1772.6 tokens and generates
373.0 decode tokens on average, recovering accuracy to 78.20\%.
Despite the additional prefill and decode cost relative to CoD,
CoD+Memory remains $1.62\times$ faster than full CoT.
These results indicate that MAC continues to provide a favorable
accuracy--latency trade-off at the 72B model scale.

\subsection{Compression Plug-in Results}
\label{app:plugin_results}

We provide the complete compression-ratio results for TokenSkip and
Extra-CoT. For TokenSkip, the reported memory-augmented variants use
top-$k=5$ Long-CoT memories. Base denotes the original compression
method without Memory, and +Mem.\ denotes its memory-augmented
counterpart. Accuracy and gains are reported in percentage points.

\subsubsection{TokenSkip}

Tables~\ref{tab:tokenskip_gsm8k_full} and
\ref{tab:tokenskip_math_full} report the complete TokenSkip results
on GSM8K and MATH-500, respectively.
All memory-augmented variants use top-$k=5$ Long-CoT memories.
Across most compression ratios and both backbone models, Memory
consistently improves the corresponding TokenSkip baseline.
The gains are generally larger under strong or intermediate
compression, while becoming smaller as the compressed baseline
approaches its uncompressed performance.

On GSM8K, Memory improves TokenSkip across all reported compression
ratios and both backbone models. On MATH-500, Memory also yields
consistent gains except for the most aggressive LLaMA-3.1-8B setting
at $\gamma=0.1$. The strongest improvements generally occur under
intermediate compression, suggesting that retrieved reasoning is most
effective when the model retains sufficient internal reasoning
capacity to interpret and apply the external scaffold.

\subsubsection{Extra-CoT}

\begin{table}[t]
\centering
\small
\setlength{\tabcolsep}{4pt}
\renewcommand{\arraystretch}{1.05}
\begin{tabular*}{\columnwidth}{
@{\extracolsep{\fill}}lrrrr
}
\toprule
Dataset & Ratio & $k$ & Base & +Mem. \\
\midrule

\multirow{5}{*}{GSM8K}
& 0.2 & 5 & 74.07 & 67.70 \\
& 0.4 & 5 & 78.92 & 71.65 \\
& 0.6 & 5 & 84.69 & 80.74 \\
& 0.8 & 5 & 85.75 & 80.52 \\
& 1.0 & 5 & 83.70 & 81.80 \\

\midrule

\multirow{5}{*}{MATH}
& 0.2 & 5 & 29.20 & 24.40 \\
& 0.4 & 5 & 31.80 & 27.00 \\
& 0.6 & 5 & 41.80 & 37.60 \\
& 0.8 & 5 & 39.80 & \textbf{41.20} \\
& 1.0 & 5 & 35.80 & \textbf{40.40} \\

\bottomrule
\end{tabular*}
\caption{
Full Extra-CoT plug-in results with Qwen2.5-7B and top-$k=5$
Long-CoT memories.
}
\label{tab:extracot_full}
\end{table}

\begin{table*}[t]
\centering
\small
\setlength{\tabcolsep}{5pt}
\renewcommand{\arraystretch}{1.10}
\begin{tabular*}{\textwidth}{
@{\extracolsep{\fill}}lrrrrrr
}
\toprule
Method
& Acc. (\%)
& \shortstack{Prefill +\\Decode (ms)}
& \shortstack{Query Tag +\\Embedding (ms)}
& \shortstack{Vector\\Search (ms)}
& \shortstack{End-to-End\\Latency (ms)}
& \shortstack{E2E\\Speedup} \\
\midrule
CoT
& 91.4
& 3742.5
& --
& --
& 3742.5
& 1.00$\times$
\\

CoD+Memory
& 89.3
& 2515.4
& 683.0
& 0.10
& 3198.5
& \textbf{1.17$\times$}
\\
\bottomrule
\end{tabular*}
\caption{
End-to-end latency breakdown on GSM8K.
Prefill + Decode denotes post-retrieval model inference latency.
For CoD+Memory, end-to-end latency additionally includes query-tag
generation, query embedding, and top-$k$ vector search.
}
\label{tab:e2e_latency_breakdown}
\end{table*}

\begin{figure*}[t]
    \centering
    \begin{minipage}{0.49\textwidth}
        \centering
        \includegraphics[width=\linewidth]{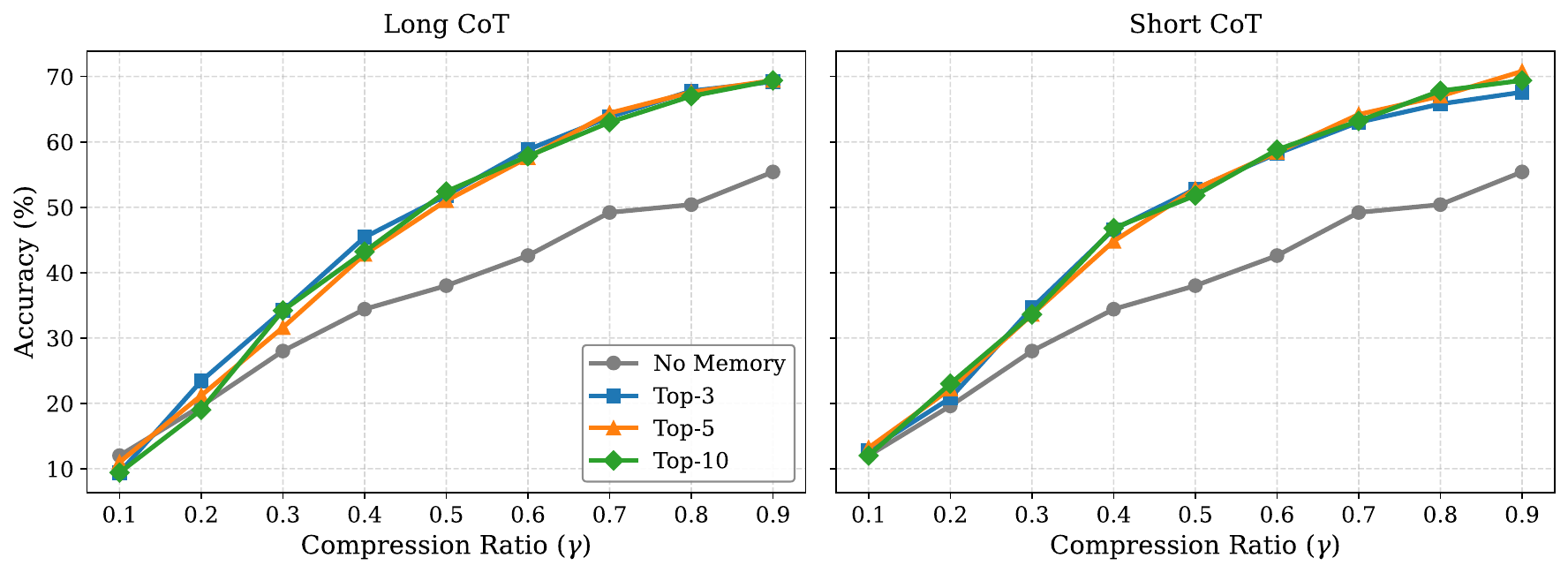}
        \small (a) Qwen2.5-7B on MATH.
    \end{minipage}
    \hfill
    \begin{minipage}{0.49\textwidth}
        \centering
        \includegraphics[width=\linewidth]{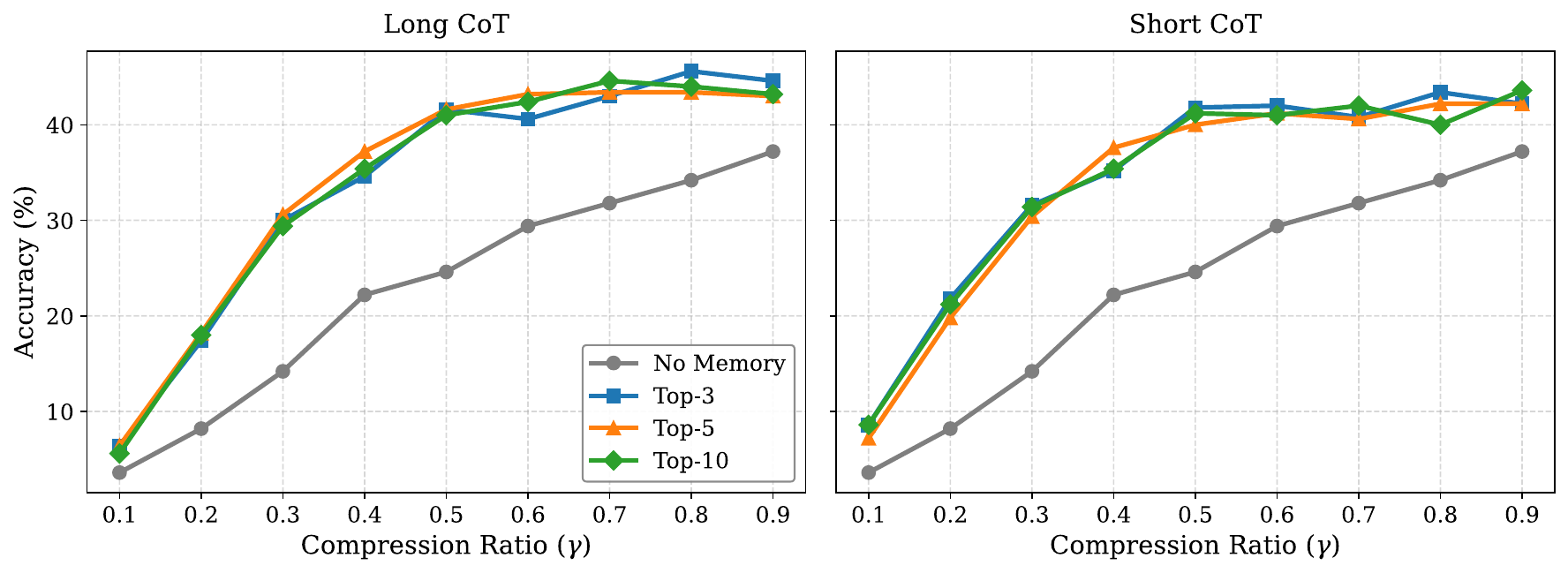}
        \small (b) LLaMA-3.1-8B on MATH.
    \end{minipage}
    \caption{
    Additional compression-ratio curves on MATH.
    We compare no-memory baselines with Memory using Long-CoT and Short-CoT formats under different retrieval sizes \(k \in \{3,5,10\}\).
    }
    \label{fig:app_math_curves}
\end{figure*}

\begin{table*}[t]
\centering
\small
\setlength{\tabcolsep}{4pt}
\renewcommand{\arraystretch}{1.08}
\begin{tabular*}{\textwidth}{
@{\extracolsep{\fill}}rrrrrrrrrrrr
}
\toprule
\multicolumn{4}{c}{GSM8K}
&
\multicolumn{4}{c}{MATH}
&
\multicolumn{4}{c}{MMLU-Sci}
\\
\cmidrule(lr){1-4}
\cmidrule(lr){5-8}
\cmidrule(lr){9-12}
$k$ & Acc. & Prefill & Decode
& $k$ & Acc. & Prefill & Decode
& $k$ & Acc. & Prefill & Decode \\
\midrule
1  & 73.39 & 259.0  & 57.9
& 1  & 52.20 & 348.8  & 152.8
& 1  & 55.07 & 267.1 & 51.6
\\
3  & 76.27 & 509.4  & 59.9
& 3  & 53.20 & 751.0  & 190.6
& 3  & 65.64 & 493.5 & 68.3
\\
5  & 77.26 & 759.8  & 61.3
& 5  & 53.40 & 1134.4 & 202.6
& 5  & 62.11 & 493.6 & 66.5
\\
8  & 78.15 & 1132.1 & 61.7
& 8  & 58.60 & 1729.8 & 218.1
& 8  & 66.52 & 493.6 & 67.7
\\
10 & 78.47 & 1379.2 & 61.9
& 10 & 58.80 & 2124.9 & 227.1
& 10 & 65.20 & 493.6 & 67.8
\\
12 & 78.62 & 1631.7 & 61.3
& 12 & 59.00 & 2522.5 & 313.7
& 12 & \textbf{66.96} & 493.5 & 68.8
\\
14 & \textbf{80.52} & 1880.9 & 61.8
& 14 & 59.60 & 2927.7 & 273.4
& 14 & 65.64 & 493.6 & 68.9
\\
16 & 79.91 & 2129.3 & 60.8
& 16 & \textbf{62.60} & 3329.5 & 297.0
& 16 & 64.32 & 493.6 & 68.3
\\
18 & 80.29 & 2378.2 & 61.5
& 18 & 62.40 & 3715.0 & 327.4
& 18 & 66.56 & 493.6 & 66.9
\\
20 & 79.68 & 2627.6 & 60.9
& 20 & 57.40 & 4120.3 & 226.5
& 20 & 65.20 & 493.6 & 69.1
\\
\bottomrule
\end{tabular*}
\caption{
Full memory-size results with Qwen2.5-7B.
We report accuracy and average prefill and decode tokens under
different values of $k$.
}
\label{tab:k_analysis_full}
\end{table*}

The Extra-CoT results reveal a method-dependent boundary of memory
augmentation. Memory improves MATH at the two highest reported ratios,
but does not improve the remaining configurations. This indicates that
retrieved reasoning is not uniformly beneficial for every compressor
and compression strength.

\subsection{End-to-End Latency Breakdown}
\label{app:latency_breakdown}

We provide a full latency breakdown to quantify the online overhead of
memory-augmented inference.
We distinguish post-retrieval model inference from complete end-to-end
latency.
Post-retrieval inference includes only model prefill and autoregressive
decoding, whereas end-to-end latency additionally includes all online
operations required to retrieve memories for the current query.

The online retrieval latency is defined as
\begin{equation}
    T_{\mathrm{ret}}
    =
    T_{\mathrm{tag}}
    +
    T_{\mathrm{emb}}
    +
    T_{\mathrm{search}},
\end{equation}
where $T_{\mathrm{tag}}$ denotes query-tag generation,
$T_{\mathrm{emb}}$ denotes query embedding, and
$T_{\mathrm{search}}$ denotes top-$k$ vector search.
The complete end-to-end latency is
\begin{equation}
    T_{\mathrm{e2e}}
    =
    T_{\mathrm{ret}}
    +
    T_{\mathrm{prefill}}
    +
    T_{\mathrm{decode}}.
\end{equation}

Table~\ref{tab:e2e_latency_breakdown} reports the latency breakdown on
GSM8K.
For standard CoT, no external retrieval is performed, and its
end-to-end latency therefore equals its prefill and decode latency.
For CoD+Memory, query-tag generation and query embedding require
$683.0$\,ms per query, while top-$k$ vector search requires only
$0.10$\,ms.
Thus, vector search itself contributes negligible overhead, whereas
query-side semantic processing constitutes the dominant retrieval cost.

After including all online retrieval operations, CoD+Memory requires
$3198.5$\,ms per query, compared with $3742.5$\,ms for standard CoT,
corresponding to a $1.17\times$ end-to-end speedup.
Therefore, although query-tag generation and embedding introduce
additional overhead, MAC still achieves lower end-to-end latency than
standard CoT. The results further suggest that optimizing query-side
representation is more important for reducing retrieval overhead than
optimizing the vector-search operation itself.

\subsection{Compression Ratio Analysis}
\label{app:compression_ratio}

To complement the GSM8K compression-ratio analysis in the main paper,
Figure~\ref{fig:app_math_curves} reports additional results on MATH.
The figure compares Long-CoT and Short-CoT memories under retrieval
sizes $k\in\{3,5,10\}$.

Across Qwen2.5-7B and LLaMA-3.1-8B, memory gains remain dependent on
compression strength. Unlike GSM8K, where the largest gains frequently
appear under strong compression, MATH often achieves its largest gains
under intermediate compression. This suggests that retrieved memories
do not fully replace model-side reasoning. On harder problems, overly
aggressive compression may remove the internal reasoning capacity
needed to interpret and apply the retrieved scaffold.

\subsection{Memory Format Analysis}
\label{app:memory_format}

Table~\ref{tab:memory_format_full} reports the complete per-$\gamma$
results for the memory-format analysis. Experiments are conducted under
TokenSkip compression on GSM8K with top-$k=5$ retrieval. The main paper
reports aggregated gains over strong, medium, and mild compression
regimes, whereas this table provides the result for each compression
ratio.

\begin{table}[t]
\centering
\small
\setlength{\tabcolsep}{4pt}
\begin{tabular*}{\columnwidth}{
@{\extracolsep{\fill}}lrrrr
}
\toprule
\multirow{2}{*}{$\gamma$}
& \multicolumn{2}{c}{LLaMA-3.1-8B}
& \multicolumn{2}{c}{Qwen2.5-7B} \\
\cmidrule(lr){2-3}
\cmidrule(lr){4-5}
& Long & Short & Long & Short \\
\midrule
0.1 & +30.71 & +31.54 & +33.66 & +29.04 \\
0.2 & +14.25 & +13.57 & +17.97 & +16.15 \\
0.3 & +7.28 & +7.05 & +11.75 & +8.49 \\
0.4 & +6.44 & +4.85 & +4.40 & +1.67 \\
0.5 & +4.70 & +3.18 & +3.56 & +1.06 \\
0.6 & +4.85 & +3.56 & +0.30 & +0.91 \\
0.7 & +1.90 & +1.44 & +0.30 & -0.45 \\
0.8 & +1.67 & +1.52 & +0.15 & -2.05 \\
0.9 & +2.58 & +0.08 & -0.45 & -0.91 \\
\bottomrule
\end{tabular*}
\caption{
Full memory-format results under TokenSkip compression on GSM8K.
Values denote accuracy gains over the corresponding no-memory
compressed baseline in percentage points.
Both Long-CoT and Short-CoT memories use top-$k=5$ retrieval.
}
\label{tab:memory_format_full}
\end{table}

The complete results follow the trend observed in the main paper:
memory gains are generally largest under stronger compression, while
Long-CoT memories remain more stable than Short-CoT memories as the
compression constraint becomes milder.

% TODO: Add Summary-memory results if Summary is presented as one of
% the primary memory formats in the method section.

\subsection{Memory Size Analysis}
\label{app:memory_size}

Table~\ref{tab:k_analysis_full} reports accuracy and token statistics
under different numbers of retrieved memories. We vary $k$ on
Qwen2.5-7B and evaluate the resulting coverage--noise trade-off.

For GSM8K and MATH, increasing $k$ substantially increases the prefill
length, while the decode length remains comparatively stable. Accuracy
initially improves as the retrieved memories provide broader reasoning
coverage, but eventually saturates or decreases as irrelevant or
redundant memories introduce additional noise. This behavior supports
the coverage--noise trade-off discussed in the main paper.

\subsection{Decoding Robustness}
\label{app:decoding_robustness}

Table~\ref{tab:temperature_full} reports the complete temperature
analysis. We evaluate full CoT and the corresponding compressed
memory-augmented method under temperatures $T\in[0,1]$, reporting
accuracy and average decode tokens.

\begin{table}[t]
\centering
\small
\setlength{\tabcolsep}{4pt}
\begin{tabular*}{\columnwidth}{
@{\extracolsep{\fill}}lrrrr
}
\toprule
Temperature
& \multicolumn{2}{c}{Standard CoT}
& \multicolumn{2}{c}{Memory} \\
\cmidrule(lr){2-3}
\cmidrule(lr){4-5}
$T$ & Acc. & Decode & Acc. & Decode \\
\midrule
0.0 & 90.1 & 282 & 81.8 & 68.2 \\
0.2 & 90.8 & 288 & 82.1 & 68.7 \\
0.4 & 91.2 & 293 & 82.3 & 69.0 \\
0.6 & 89.8 & 297 & 81.9 & 68.5 \\
0.8 & 90.5 & 301 & 82.0 & 68.9 \\
1.0 & 91.0 & 304 & 82.0 & 69.1 \\
\midrule
Range & 1.40 & 22.00 & 0.50 & 0.90 \\
Across-$T$ Std. & 0.49 & 7.51 & 0.16 & 0.31 \\
\bottomrule
\end{tabular*}
\caption{
Full temperature results for decoding robustness.
Accuracy is reported in percentage points, and Decode denotes the
average number of generated tokens.
Across-$T$ Std.\ denotes the standard deviation across the six
temperature settings rather than across repeated random runs.
}
\label{tab:temperature_full}
\end{table}

Across the evaluated temperatures, the memory-augmented method exhibits
smaller variation in both accuracy and decode length. However, this
analysis measures sensitivity across temperature settings and should
not be interpreted as a repeated-seed estimate of statistical
uncertainty.

\section{Statement on the Use of AI Assistants}
\label{sec:appendix_ai}
During the preparation of this work, we utilized an AI assistant (e.g., ChatGPT) for tasks such as proofreading and improving language clarity. The core ideas, experimental design, and final text were authored by the human authors, who take full responsibility for the content of this paper.

\section{Data and Code Availability}
\label{app:data_code_availability}

All datasets used in this work are publicly available academic benchmarks,
including GSM8K~\citep{cobbe2021training},
MATH~\citep{hendrycks2021measuring},
BBH~\citep{suzgun2023challenging},
MMLU Science~\citep{hendrycks2020measuring}, and AIME 2024.
We use these datasets solely for research purposes and follow their
respective licenses and terms of use.

For baseline methods, including TokenSkip, RPC, and Extra-CoT, we use
the officially released implementations whenever available or reproduce
them according to their reported settings. An anonymized code package
containing the MAC implementation, experimental configurations, and
evaluation utilities is included in the supplementary material.
Upon publication, we will release the source code under the MIT License. 

\end{document}